\documentclass{sig-alternate-05-2015}
\usepackage{amsmath}
\usepackage{array}
\newtheorem{theorem}{Theorem}
\usepackage{xcolor}
\usepackage{amssymb}
\usepackage{amsfonts}
\usepackage[mathscr]{euscript}
\usepackage{mathtools}
\usepackage{graphicx}	
\usepackage{caption}
\usepackage{multirow}
\usepackage{hyperref}
\usepackage{subcaption}
\usepackage{algorithm}
\usepackage{algpseudocode}
\usepackage{array}
\usepackage{pdflscape}
\makeatletter
\newcolumntype{P}[1]{>{\centering\arraybackslash}p{#1}}
\newcommand{\thickhline}{%
    \noalign {\ifnum 0=`}\fi \hrule height 1pt
    \futurelet \reserved@a \@xhline
}
\newcolumntype{"}{@{\hskip\tabcolsep\vrule width 1pt\hskip\tabcolsep}}

\begin{document}
\setcopyright{acmcopyright}

\doi{10.475/123_4}

\isbn{123-4567-24-567/08/06}

\conferenceinfo{KDD '16}{August 13--17, 2016, San Francisco, CA, USA}

\acmPrice{\$15.00}

\title{Lexis: An Optimization Framework for Discovering \\the Hierarchical Structure of Sequential Data}

\numberofauthors{3}
\author{
\alignauthor
Payam Siyari\\
       \affaddr{GeorgiaTech}\\
       \affaddr{Atlanta, GA, USA}\\
       \email{payamsiyari@gatech.edu}
\alignauthor
Bistra Dilkina\\
       \affaddr{GeorgiaTech}\\
       \affaddr{Atlanta, GA, USA}\\
       \email{bdilkina@cc.gatech.edu}
\alignauthor
Constantine Dovrolis\\
       \affaddr{GeorgiaTech}\\
       \affaddr{Atlanta, GA, USA}\\
       \email{constantine@gatech.edu}
}

\CopyrightYear{2016} 
\setcopyright{acmcopyright}
\conferenceinfo{KDD '16,}{August 13-17, 2016, San Francisco, CA, USA}
\isbn{978-1-4503-4232-2/16/08}\acmPrice{\$15.00}
\doi{http://dx.doi.org/10.1145/2939672.2939741}

\maketitle

\begin{abstract}
Data represented as strings abounds in biology, linguistics, document mining, web search and many other fields.
Such data often have a hierarchical structure, either because they were artificially designed and composed in a hierarchical manner or because there is an underlying evolutionary process that creates repeatedly more complex strings from simpler substrings.
We propose a framework, referred to as {\em Lexis}, that produces an optimized 
hierarchical representation of a given set of \lq\lq target\rq\rq~ strings. 
The resulting hierarchy, \lq\lq Lexis-DAG\rq\rq, shows how to construct each target through 
the concatenation of intermediate substrings, minimizing the total number of such concatenations or DAG edges. 
The Lexis optimization problem is related to the smallest grammar problem. 
After we prove its NP-hardness for two cost formulations, 
we propose an efficient greedy algorithm for the construction of Lexis-DAGs. 
We also consider the problem of identifying the set of intermediate nodes (substrings) 
that collectively form the \lq\lq core\rq\rq~ of a Lexis-DAG, which is important in the analysis of Lexis-DAGs.
We show that the Lexis framework can be applied in diverse applications such as 
optimized synthesis of DNA fragments in genomic libraries, 
hierarchical structure discovery in protein sequences,
dictionary-based text compression, and feature extraction from a set of documents.

\end{abstract}

%
\section{Introduction}

In both nature and technology, information is often represented in sequential form, 
as strings of characters from a given alphabet \cite{stringsBook}. 
Such data often exhibit a hierarchical structure in which previously constructed 
strings are re-used in composing longer strings \cite{sequitur}.
In some cases this hierarchy is formed ``by design'' in synthetic processes where there are some cost savings
associated with the re-use of existing modules \cite{kleinberg,costHierarchy}.
In other cases, the hierarchy
emerges naturally when there is an underlying evolutionary process that repeatedly creates
more complex strings from simpler ones, conserving only those that are being re-used 
\cite{costHierarchy,sequitur}. 
For instance, language is hierarchically organized starting from phonemes to stems, words, compound words,
phrases, and so on \cite{langHierarchy}. 
In the biological world, genetic information is also represented sequentially and there is ample evidence
that evolution has led to a hierarchical structure in which sequences of DNA bases are first translated into 
amino acids, then form motifs, regions, domains, and this process continues to 
create many thousands of distinct proteins \cite{dubitzky2007fundamentals}.  

In the context of synthetic design, an important problem is to construct a minimum-cost Directed Acyclic
Graph (DAG) that shows how to produce a given set of ``target strings'' from a given ``alphabet'' in a hierarchical
manner, through the construction of intermediate substrings that are re-used in at least two higher-level 
strings. The cost of a DAG should be related somehow to the amount of ``concatenation work'' (to
be defined more precisely in the next setion) that the corresponding hierarchy would require.   
For instance, in \emph{de novo} DNA synthesis \cite{dnaSynthPaper, dnaSynthPaperNew}, 
biologists aim to construct target DNA sequences by concatenating previously synthesized 
DNAs in the most cost-efficient manner. 

In other contexts, 
it may be that the target strings were previously constructed through an evolutionary process (not 
necessarily biological), or that the
synthetic process that was followed to create the targets is unknown.
Our main premise is that even in those cases it is still useful to construct a cost-minimizing DAG that composes 
the given set of targets hierarchically, through the creation of intermediate substrings.
The resulting DAG shows the most parsimonious way to represent the given targets hierarchically,
revealing substrings of different lengths that are highly re-used in the targets and identifying the
dependencies between the re-used substrings.
Even though it would not be possible to prove that the given targets were actually constructed through the
inferred DAG, this optimized DAG can be thought of as a plausible hypothesis for the unknown process that created
the given targets as long as we have reasons to believe that that process cares to minimize, even heuristically,
the same cost function that the DAG optimization considers. 
Additionally, even if our goal is not to reverse engineer the process that generated the given targets,
the derived DAG can have practical value in the applications such as compression or feature extraction.

In this paper, we propose an optimization framework, 
referred to as {\em Lexis},\footnote{Lexis means ``word'' in Greek.}
that designs a minimum-cost 
hierarchical representation of a given set of target strings.
The resulting hierarchy, referred to as ``Lexis-DAG'', shows how to construct each target through
the concatenation of intermediate substrings, which themselves might be the result of concatenation of other shorter substrings, all the way to a given alphabet of elementary symbols.
We consider two cost functions: minimizing the total number of concatenations and minimizing the 
number of DAG edges. The choice of cost function is application-specific. 
The Lexis optimization problem is related to the smallest grammar problem \cite{sgp, gcodes}.
We show that Lexis is NP-hard for both cost functions,
and propose an efficient greedy algorithm for the construction of Lexis-DAGs.
Interestingly, the same algorithm can be used for both cost functions.
We also consider the problem of identifying the set of intermediate nodes (substrings)
that collectively form the ``core'' of a Lexis-DAG. 
This is the minimal set of DAG nodes that can cover a given fraction of source-to-target paths,
from alphabet symbols to target strings. 
The core of a Lexis-DAG represents the most central substrings in the corresponding hierarchy.
We show that the Lexis framework can be applied in diverse applications such as
optimized synthesis of DNA fragments in genomic libraries,
hierarchical structure discovery in protein sequences,
dictionary-based text compression, and feature extraction from a set of documents.

\section{Problem Statement}
\subsection{Lexis-DAG}
Given an alphabet $S$ and a set of ``target'' strings $T$ over the alphabet $S$, we need to construct a Lexis-DAG.
A Lexis-DAG $D$ is a directed acyclic graph $D(V,E)$, where $V$ is the set of nodes and $E$ the set of edges, 
that satisfies the following three constraints.\footnote{To simplify the notation, even though $D$ is 
a function of $S$ and $T$, we do not denote it as such.} 

First, each node $v \in V$ in a Lexis-DAG represents a string $\mathcal{S}(v)$
of characters from the alphabet $S$.
The nodes $V_S$ that represent characters of $S$ are referred to as \emph{sources}, and they have zero in-degree.
The nodes $V_T$ that represent target strings $T=\{t_1, t_2, \dots, t_m\}$ are referred to as {\em targets}, and they
have zero out-degree.
$V$ also includes a set of {\em intermediate nodes} $V_M$, which represent substrings that appear in the targets $T$.  
So, $V=V_S\cup V_M\cup V_T$.

Second,
each node in $V_M\cup V_T$ of a Lexis-DAG represents a string that is the concatenation 
of two or more substrings, specified by the incoming edges from other nodes to that node. 
Specifically, 
an edge $e \in E$ from node $u$ to node $v$ is a triplet $(u,v,i)$ such that
the string $\mathcal{S}(u)$ appears as substring of $\mathcal{S}(v)$ at index $i$ 
(the first character of a string has index 1).
Note that there may be more than one edges from node $u$ to node $v$. 
The number of incoming and outgoing edges for $v$ is denoted by $d_{in}(v)$ and $d_{out}(v)$, respectively. 
$I(v)$ is the sequence of nodes $u$ that appear in the incoming edges $(u,v,i)$ of $v$, ordered by edge index $i$. 
We require that for each node $v$ in $V_M\cup V_T$ replacing the sequence of nodes in $I(v)$ with their corresponding strings results in exactly $\mathcal{S}(v)$.

\begin{figure}
\center
\begin{subfigure}[b]{0.15\textwidth}
\includegraphics[trim = 1.3cm 1.35cm 1.40cm 1.39cm,clip,scale=0.37]{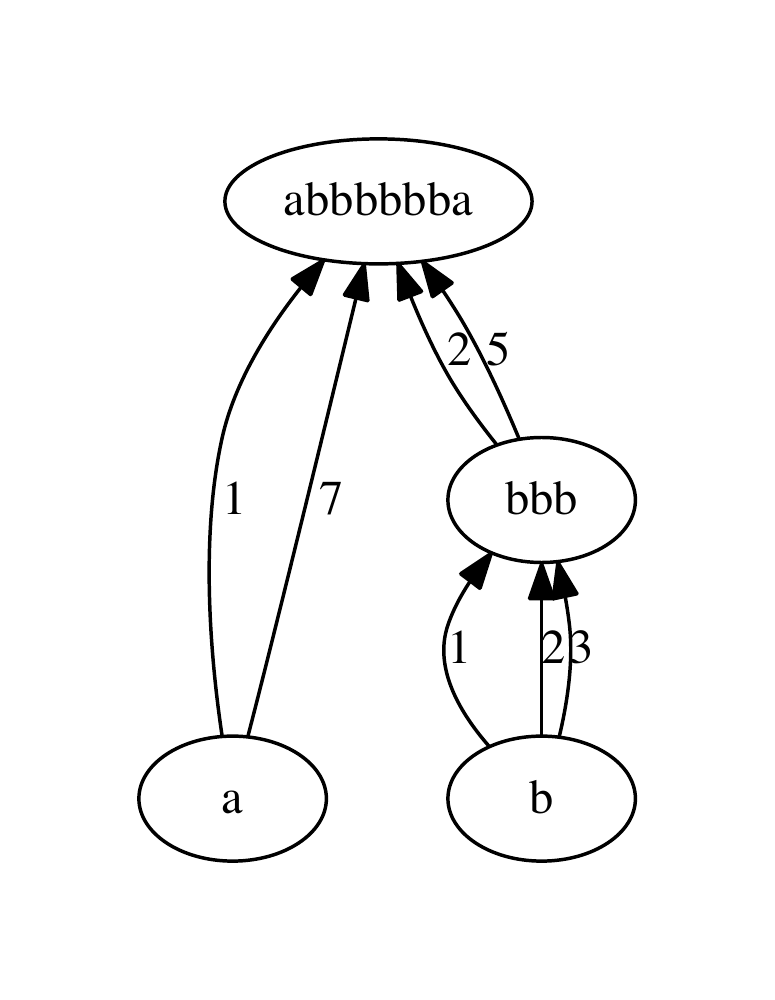}
\caption{~}
\end{subfigure}
\hfill{
\begin{subfigure}[b]{0.15\textwidth}
\includegraphics[trim = 1.4cm 1.35cm 1.35cm 1.39cm,clip,scale=0.37]{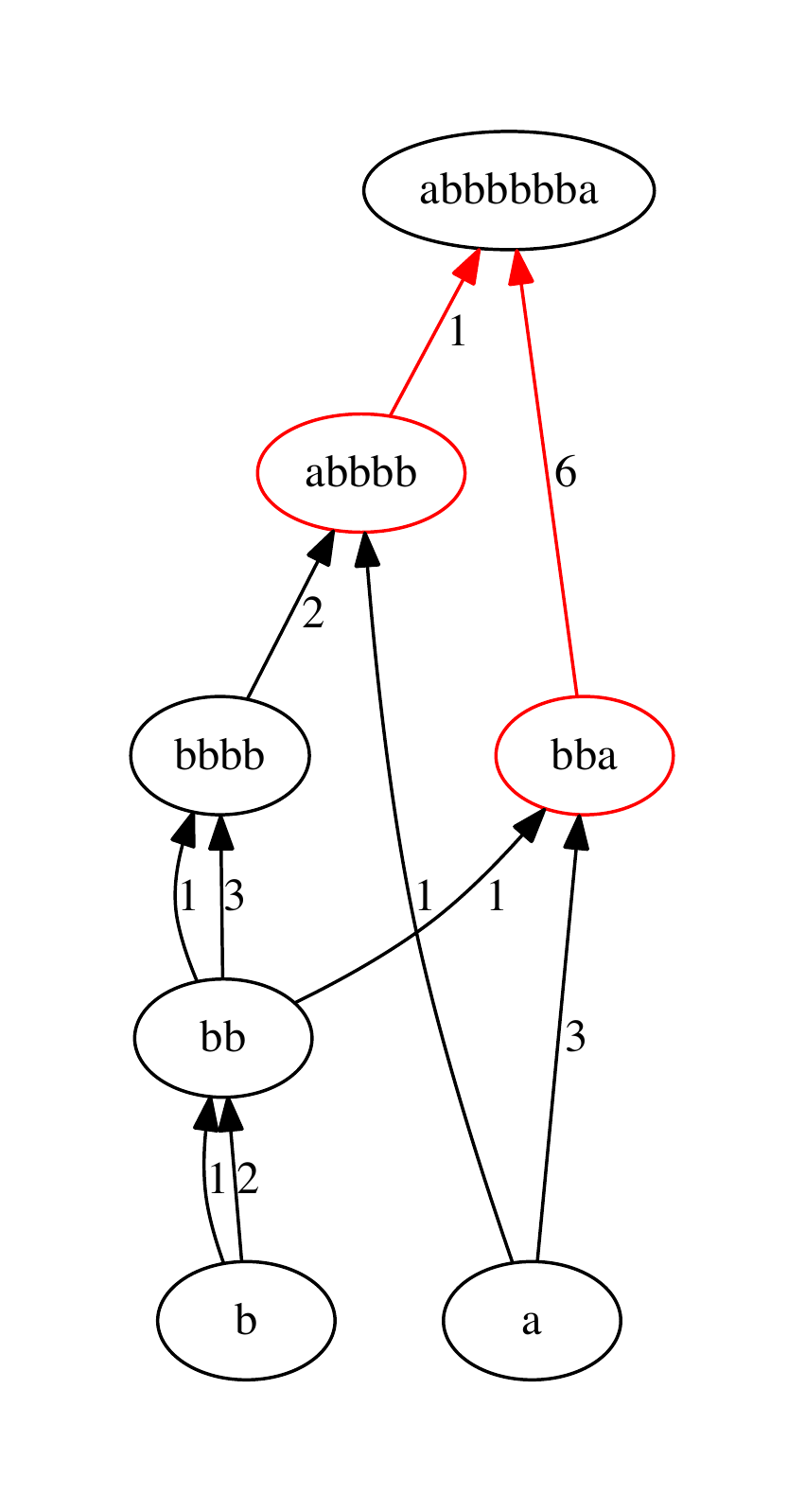}
\caption{~}
\end{subfigure}
\hfill{
\begin{subfigure}[b]{0.15\textwidth}
\includegraphics[trim = 1.3cm 1.35cm 1.35cm 1.39cm,clip,scale=0.37]{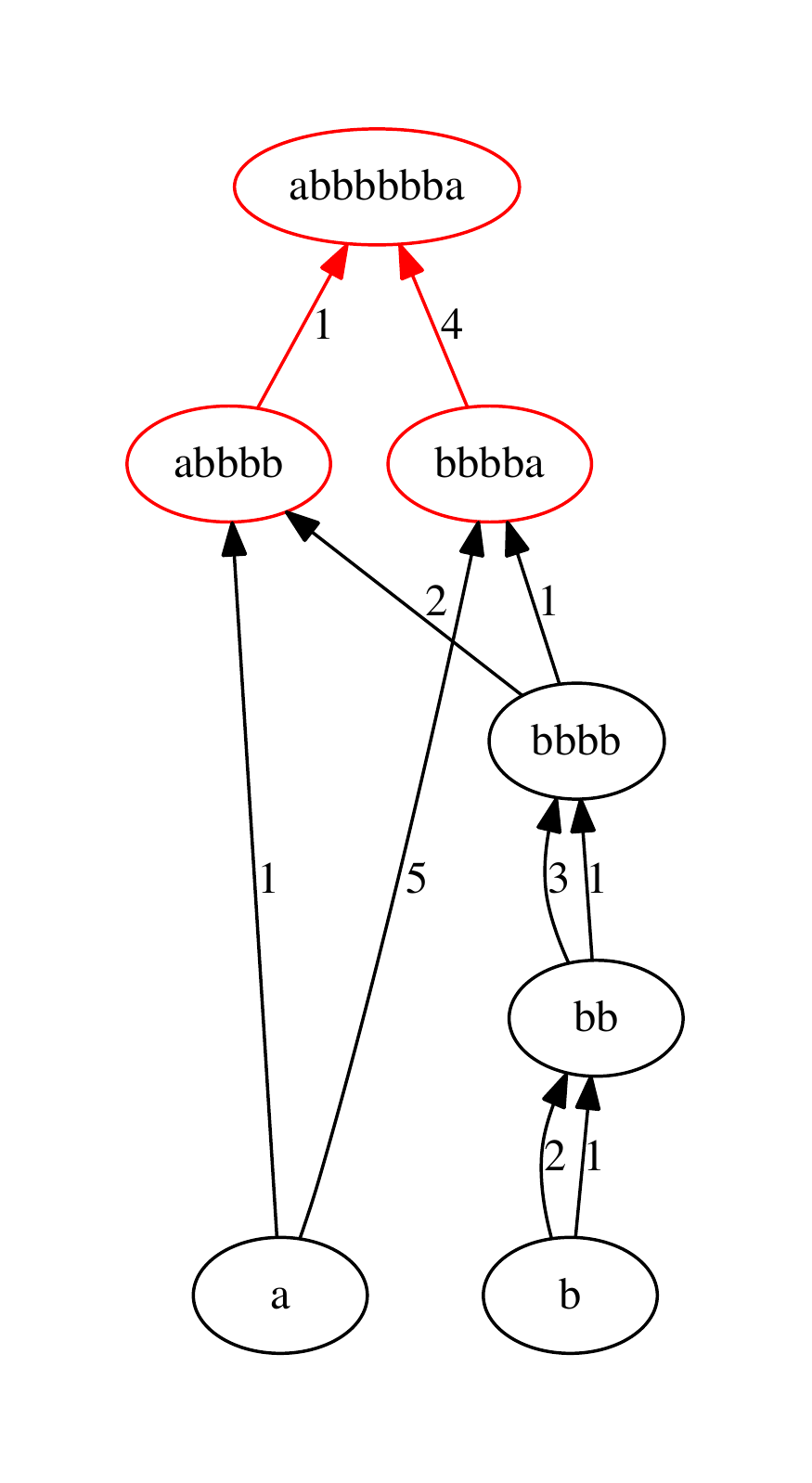}
\caption{~}
\end{subfigure}}}
\vspace{-3.5mm}
\caption{\small Illustration of the Lexis-DAG  for targets $T=\{abbbbbba\}$ and sources $S=\{a,b\}$. Edge-labels indicate the occurrence indices: {\bf (a)} A valid Lexis-DAG having both minimum number of concatenations and edges. {\bf (b)} An invalid Lexis-DAG: two intermediate nodes are re-used only once. {\bf (c)} An invalid Lexis-DAG: the top-layer string is not equal to the concatenation of its two in-neighbors (best viewed in color).}
\label{dagDef}
\vspace{-5mm}
\end{figure}

Third, a Lexis-DAG should only include intermediate nodes that have an out-degree of at least two,
\begin{equation}
\forall v\in V_M, d_{out}(v) \geq 2. \label{min-dout-constraint}
\end{equation}
In other words, every intermediate node $v \in V_M$ in a Lexis-DAG should be such that 
the string $\mathcal{S}(v)$ is re-used in at least two concatenation operations. 
Otherwise, $\mathcal{S}(v)$ is either not used in any concatenation operation, 
or it is used only once and so the outgoing 
edge from $v$ can be replaced 
by re-wiring the incoming edges of $v$ straight to the single occurrence of $\mathcal{S}(v)$. 
In both cases node $v$ can be removed from the Lexis-DAG, resulting in a more parsimonious
hierarchical representation of the targets.  
%
Fig.~\ref{dagDef} illustrates the concepts introduced in this section.

\subsection{The Lexis Optimization Problem}
The {\em Lexis} optimization problem is to construct a minimum-cost Lexis-DAG for the 
given alphabet $S$ and target strings $T$.
In other words, the problem is to determine the set of intermediate nodes $V_M$ and all
required edges $E$ so that the corresponding Lexis-DAG $D$ is optimal in terms of a given cost
function $C(D)$.  

\begin{equation}
\label{Lexis}
\begin{aligned}
\text{~}&\min_{(E,V_{M})}\text{~}C(D)\\
\text{~}&s.t.\text{~} D=(V,E)\text{ is a Lexis-DAG for $S$ and $T$}
\end{aligned}
\end{equation}

The selection of an appropriate cost function is somewhat application-specific. 
A natural cost function to consider is the number of edges in the Lexis-DAG.
In certain applications, such as DNA synthesis, the cost is usually measured in terms of the
number of required concatenation operations. 
In the following, we consider both cost functions. 
Note that we choose to not explicitly minimize the number of intermediate nodes in $V_M$;
minimizing the number of edges or concatenations, however, tends to also reduce the number of 
required intermediate nodes.  
Additionally, the constraint (\ref{min-dout-constraint}) means that the optimal Lexis-DAG will
not have redundant intermediate nodes that can be easily removed without increasing the 
concatenation or edge cost. 
More general cost formulations, such as a variable edge cost or a weighted average 
of a node cost and an edge cost, are interesting but they are not pursued in this paper.

\subsubsection{Edge cost}
Suppose that the cost of each edge is one. 
The {\em edge cost} to construct a node $v \in V$ is defined as the number of incoming edges required
to construct $\mathcal{S}(v)$ from its in-neighbors, which is equal to $d_{in}(v)$. 
The edge cost of source nodes is obviously zero. 
The edge cost $\mathcal{E}(D)$ of Lexis-DAG $D$ is defined as the edge cost of all nodes, equal to the number of edges in $D$: 

\vspace{-5mm}
\begin{equation}
\mathcal{E}(D) = \sum_{v\in V} d_{in}(v) = |E|
\label{firstDAGCost}
\end{equation}

%

\vspace{-3mm}
\subsubsection{Concatenation cost}
Suppose that the cost of each concatenation operation is one. 
The {\em concatenation cost} to construct a node $v \in V_M \cup V_T$ 
is defined as the number of concatenations required
to construct $\mathcal{S}(v)$ from its in-neighbors, which is equal to $d_{in}(v)-1$. 
The concatenation cost $\mathcal{C}(D)$ of Lexis-DAG $D$ is defined as the concatenation cost of all 
non-source nodes;
it is easy to see that this is equal to the number of edges in $D$ minus the number of non-source nodes, 
\begin{multline}
\mathcal{C}(D) = \sum_{v\in V \backslash V_S}\left(d_{in}(v)-1\right) = |E|-|V \backslash V_S|
\label{secondDAGCost}
\end{multline}

%
%

\begin{theorem}
The optimization problem in Eq. \eqref{Lexis} is NP-hard for both the edge cost of Eq. \eqref{firstDAGCost} and the concatenation cost of Eq. \eqref{secondDAGCost}.
\label{nphardness:both}
\end{theorem}
The proof is given in the Appendix. Note that the objective in Eq.~\eqref{secondDAGCost} is an explicit function of the number of intermediate
nodes in the Lexis-DAG. 
Hence the optimal solutions for the concatenation cost can be different than 
those for the edge cost.
An example is shown in the Appendix.

\vspace{-2mm}
\section{The Greedy Lexis algorithm}

\begin{figure}[t]
\begin{tabular}{m{4cm} m{5cm}}
\\
\includegraphics[trim = 1.3cm 1.4cm 1.44cm 1.4cm,clip,scale=0.38]{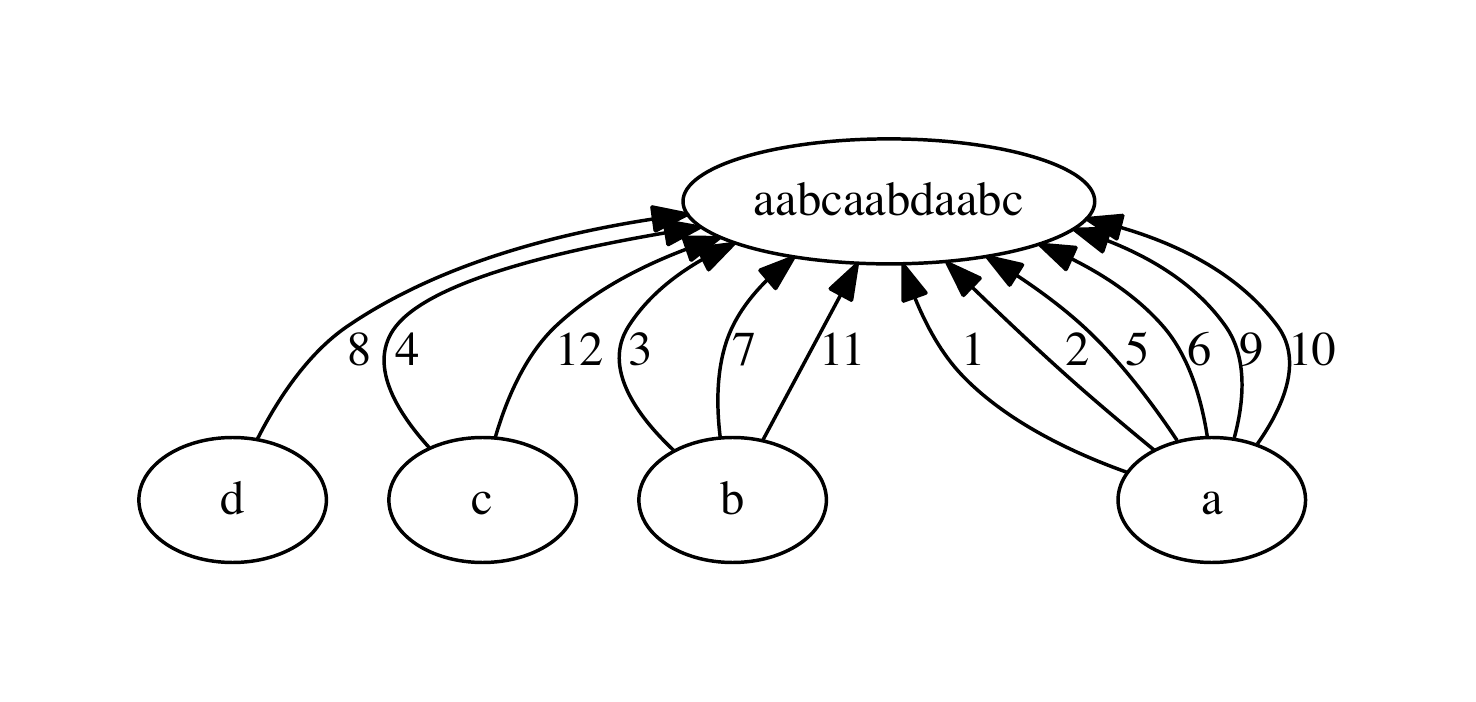}
&\multirow{2}{*}{\includegraphics[trim = 1.4cm 1.35cm 1.35cm 1.39cm,clip,scale=0.42]{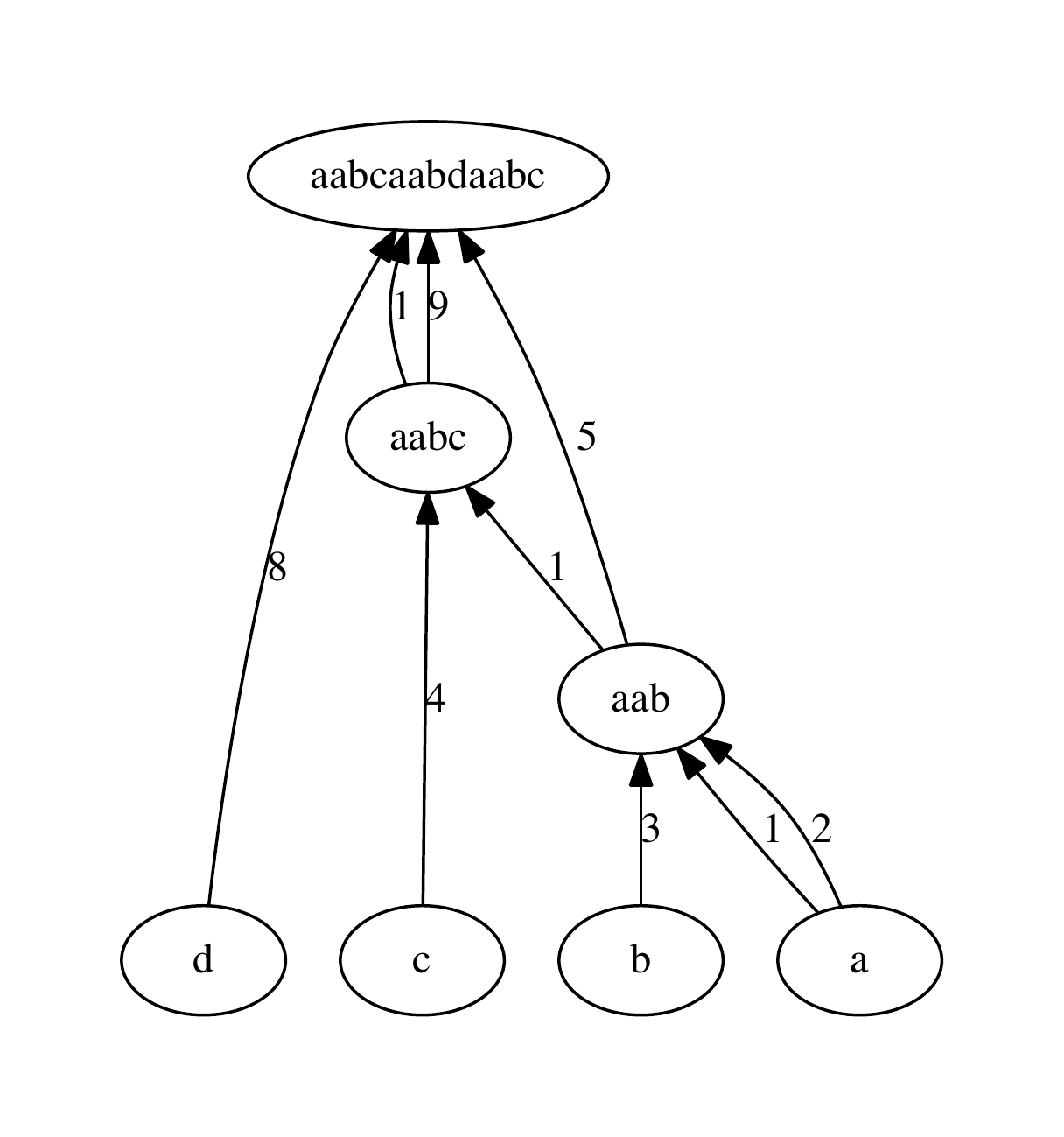}}\\
$~~~~~~~~~~~~~~~~~~~~~~~$(a)
\\\\
\includegraphics[trim = 1.37cm 1.4cm 1.4cm 1.39cm,clip,scale=0.4]{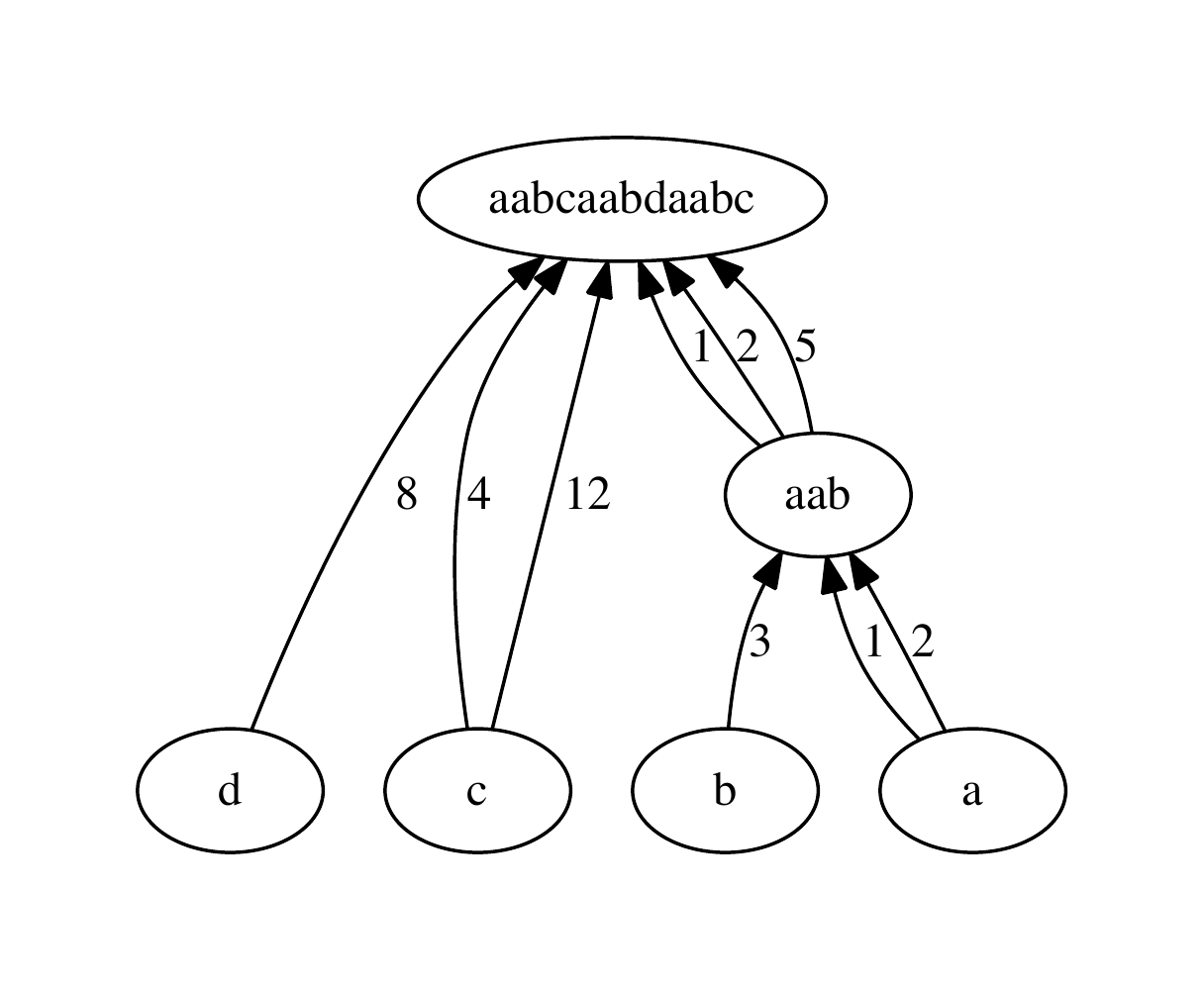}
\\
$~~~~~~~~~~~~~~~~$(b) & $~~~~~~~~~~~~~~~~$(c)
\end{tabular}
\caption{\small Illustration of \textsc{G-Lexis} given target $T=\{aabcaabdaabc\}$ and sources $S=\{a,b,c,d\}$. - {\bf (a)} Initial Lexis-DAG: The string passed to the suffix tree is $I(aabcaabdaabc)=aabcaabdaabc$. {\bf (b)} Substring \emph{aab} has maximum {\em SavedCost}. The target can be now written as $I(aabcaabdaabc)=\sigma_{_{aab}}c\sigma_{_{aab}}d\sigma_{_{aab}}c$ where $\sigma_{_{aab}}$ is the substring \emph{aab} of the new intermediate node. We also have that $I(aab)=aab$. The strings passed to the suffix tree are $\{I(aabcaabdaabc),I(aab)\}$. {\bf (c)} Substring $\sigma_{_{aab}}c$ has maximum {\em SavedCost} and is chosen for a new intermediate node. In this example, this would be the last iteration.}
\label{glexisExample}
\vspace{-5mm}
\end{figure}


In this section, we describe a greedy algorithm, referred to as
 \textsc{G-Lexis}, for both previous optimization problems.
The basic idea in \textsc{G-Lexis} is that it searches for the substring $\xi$
that will lead, under certain assumptions, to the maximum cost reduction
when added as a new intermediate node in the Lexis-DAG. 
The algorithm starts from the trivial Lexis-DAG with no intermediate nodes and edges from the source nodes representing alphabet symbols to each of their occurrences in the target strings.

Recall that for every node $v\in V_T\cup V_M$, $I(v)$ is the sequence of nodes appearing in the incoming edges of $v$, 
i.e., the sequence of nodes whose string concatenation results in the string $\mathcal{S}(v)$ represented by $v$.
The sequences $I(v)$ can be interpreted as strings over the ``alphabet'' of Lexis-DAG nodes. 
Note that every symbol in a string $I(v)$ has a corresponding edge in the Lexis-DAG.
We look for a repeated substring $\xi$ in the strings $I_{T\cup M} = \{I(v) | v\in V_T\cup V_M\}$ that can be used to construct a new intermediate node.
We can construct a new intermediate node for $\xi$, 
create incoming edges based on the symbols in $\xi$
(remember $\xi$ is a substring over the alphabet of nodes),
and replace the incoming edges to each of the non-overlapping repeated occurrences of $\xi$ with a single outgoing edge from the new node.

Consider the edge cost first.
Suppose that $\xi$ is repeated $R_{T\cup M,\xi}$ times in the strings
$I_{T\cup M}$. 
If these occurrences of $\xi$ are non-overlapping, the number of required edges would be  
$|\xi| \, R_{T\cup M,\xi}$.
After we construct a new intermediate node for $\xi$ as outlined above, 
the edge cost will be $|\xi| + R_{T\cup M,\xi}$.
So, the reduction in edge cost from re-using $\xi$ would be  
$(R_{T\cup M,\xi}-1)(|\xi|-1) -1$.
Under the stated assumptions about $\xi$,
this reduction is non-negative if $\xi$ is repeated at least twice and its length is at least two. 

Consider the concatenation cost now. 
If these occurrences of $\xi$ are non-overlapping, the number of required concatenations for all the repeated occurrences would be  
$(|\xi| - 1) \, R_{T\cup M,\xi}$.
After we construct a new intermediate node for $\xi$ as outlined above, 
the concatenation cost will be $|\xi| -1$.
We expect a reduction in the number of required concatenations by 
$(R_{T\cup M,\xi}-1) (|\xi|-1)$.

So, the greedy choice for both cost functions is the same: select the substring $\xi$ 
that maximizes the term $\mathit{SavedCost}=(R_{T\cup M,\xi}-1) (|\xi|-1)$.
For this reason, our \textsc{G-Lexis} algorithm can be used for both cost functions we consider. 
It starts with the trivial Lexis-DAG, and at each iteration it chooses a substring of $I_{T\cup M}$ in the Lexis-DAG that maximizes {\em SavedCost}, creates a new intermediate node for that substring and updates the edges of the Lexis-DAG accordingly.
The algorithm terminates when there are no more substrings of $I_{T\cup M}$ with length at least two and repeated at least twice. 
The pseudocode for \textsc{G-Lexis} is shown in Algorithm~1. 
An example of application of the \textsc{G-Lexis} algorithm is shown in Fig. \ref{glexisExample}.

\begin{algorithm}
\textsc{Algorithm 1. G-Lexis}
\small
{~\\\linethickness{0.5mm}\line(1,0){243}}
{\bf Input:}~\text{Alphabet $S$, Targets $T$}
{\bf Output:}~\text{Lexis-DAG $D$}\vspace{-2mm}
{~\\\linethickness{0.1mm}\line(1,0){243}}
\begin{enumerate}
\item
Initialize $V \leftarrow V_T \cup V_S$ and $E$, constructing each target in $T$ from characters in $S$.
$V_M \leftarrow \varnothing$. 
\vspace{-1.5mm}
\item
Repeat:
\vspace{-1.5mm}
\begin{enumerate}
\item
$I_{T\cup M}$\@$\leftarrow$\@ $\{I(v) | v \in (V_T\cup V_M)\}$;
\item
Select $\xi$ with maximum $(R_{T\cup M,\xi}-1)(|\xi|-1)$, where $R_{T\cup M,\xi}$ is the number of repeats of substring $\xi$ in $I_{T\cup M}$; \url{\#Greedy Choice}
\item 
If $(R_{T\cup M,\xi}-1)(|\xi|-1) = 0$, \emph{break;}
\item
$V$\@$\leftarrow$\@$V \cup \{\sigma_{\xi}\}$, where $\sigma_{\xi}$ is the new intermediate node, and update $E$ accordingly;$~~$\url{\#Update Lexis-DAG}
\end{enumerate}
\vspace{-3mm}
\end{enumerate}
\label{glexis}
\end{algorithm}

%


At each iteration of \textsc{G-Lexis}, we need to find efficiently the substring of $I_{T\cup M}$ with maximum {\em SavedCost}. 
We observe that the substring that maximizes {\em SavedCost} is a ``maximal repeat''. 
Maximal repeats are substrings of length at least two, whose extension to the right or 
left would reduce its occurrences in the given set of strings. 
Suppose that it is not. 
Then, there is a substring $\hat{\sigma}$, which is not a maximal repeat, that maximizes {\em SavedCost}.
If we can extend $\hat{\sigma}$ to the left or right we can increase its length 
without reducing its number of occurrences.
By doing so, we construct a new substring with higher {\em SavedCost} than $\hat{\sigma}$,
violating our initial assumption.
So, the substring that maximizes {\em SavedCost} is a maximal repeat.
A suffix tree over a set of input strings captures all right-maximal repeats, and right-maximal repeats are a superset of all maximal repeats \cite{stringsBook}. 
To pick the one with maximum {\em SavedCost}, 
we need the count of non-overlapping occurrences of these substrings. 
A Minimal Augmented Suffix Tree~\cite{minAug} over $I_{T\cup M}$ can be constructed and used to count the number of non-overlapping occurrences of all right-maximal repeats in overall $O(L \log L)$ time, where $L$ is the total length of target strings. 
Using a regular suffix tree instead, this can be achieved in only $O(L)$ time; 
but suffix tree may count overlapping occurrences.
In our implementation we prefer to use regular suffix tree, following related work  \cite{galle} that has shown 
that this performance optimization has negligible impact on the solution's quality.
So, the substring that is chosen for the new Lexis-DAG node is based on length and overlapping occurrence count. 
We then use the suffix tree to iterate over all occurrences of the selected substring, 
skipping overlapping occurrences. 
If a selected substring has less than two non-overlapping occurrences, we skip to the next best substring.
Using the suffix tree, we can update the Lexis-DAG with the new intermediate node, and with 
the corresponding edges for all occurrences of that substring, in $O(L)$ time. 
The maximum number of iterations of \textsc{G-Lexis} is $O(L)$ because each iteration reduces the number of edges (or concatenations), which at the start is $O(L)$. 
So, the overall run-time complexity using suffix tree is $O(L^2)$.

We have also experimented with other algorithms, such as a greedy heuristic that selects the longest repeat in each iteration of building the DAG, i.e., it chooses based on length among all substrings that appear at least twice in the targets or intermediate node strings.
This heuristic can be efficiently implemented to run in only $O(L)$ time \cite{linearlnrs}.
Our evaluation shows that \textsc{G-Lexis} performs significantly better than the longest repeat heuristic in terms of solution quality, 
despite some running time overhead. 
Running both algorithms on a machine with an Intel Core-i7 2.9 GHz CPU and 16GB of RAM 
on the NSF abstracts dataset (introduced in Section 5) of $2,309$ target strings with total length $245,968$ symbols
takes 562~sec for \textsc{G-Lexis} and 408~sec for the longest repeat algorithm.
The edge cost with \textsc{G-Lexis} is 169,060 compared to 183,961 with the longest repeat algorithm. 
More detailed results can be found in the Appendix.

\vspace{-2mm}
\section{Path-Centrality and the Core of a Lexis-DAG}
After constructing a Lexis-DAG, an important question is to rank the constructed intermediate
nodes in terms of significance or {\em centrality.}
Even though there are many related metrics in the network analysis literature, 
such as closeness, betweenness or eigenvector centrality \cite{network-science-book-Newman},  
none of them captures well the semantics of a Lexis-DAG.
In a Lexis-DAG, a path that starts from a source and terminates at a target represents a dependency chain 
in which each node depends on all previous nodes in that path.
So, the higher the number of such source-to-target paths traversing an intermediate node $v$ is,
the more important $v$ is in terms of the number of dependency chains it participates in.
More formally, let $P_D(v)$ be the number of source-to-target paths that traverse node $v \in V_M$;
we refer to $P_D(v)$ as the {\em path centrality} of intermediate node $v$.
The path centrality of sources and targets is zero by definition. First, note that: 
\begin{equation}
\vspace{-2mm}
P(v) = P_S(v) \, P_T(v)
\end{equation}
where $P_S(v)$  is the number of paths from any source to $v$, 
and $P_T(v)$ is the number of paths from $v$ to any target.
This suggests an efficient way to calculate the path centrality of all nodes in a Lexis-DAG in $O(|E|)$ time:
perform two DFS traversals, one starting from sources and following the direction of edges,
and another starting from targets and following the opposite direction.  
The first DFS traversal will recursively produce $P_S(v)$
while the second will produce $P_T(v)$, for all intermediate nodes.

Second, it is easy to see that $P_T(v)$ 
is equal to the number of times string $\mathcal{S}(v)$ is used for replacement in the target strings $T$.
Similarly, $P_S(v)$ is equal  to the number of times any source node is repeated in 
$\mathcal{S}(v)$, which is simply the length of $\mathcal{S}(v)$. 
So, the path centrality $P(v)$ of a node in a Lexis-DAG can be also interpreted as its ``re-use count''
(or number of replaced occurrences in the targets) times its length.
Thus, an intermediate node will rank highly in terms of path centrality if it is 
both long and frequently re-used.

An important follow-up question is to identify the {\em core} of a Lexis-DAG, i.e., a set 
of intermediate nodes that represent, as a whole, the most important substrings in that Lexis-DAG. 
Intuitively, we expect that the core should include nodes of high path centrality, and that almost 
all source-to-target dependency chains of the Lexis-DAG should traverse at least one of these core nodes. 

More formally, suppose $K$ is a set of intermediate nodes 
and $\mathscr{P}^-(K)$ is the set of source-to-target paths after we remove the nodes in $K$ from $D$. 
The core of $D$ is defined as the minimum-cardinality set of intermediate nodes $\hat{K}$ such that
the fraction of remaining source-to-target paths after the removal of $\hat{K}$ is at most $\tau$: 
\begin{equation}
\vspace{-2mm}
\label{coreid}
\begin{aligned}
\min_{\text{~}K \subseteq V_M}\text{~~~}&|K|\\
s.t.\text{~}&|\mathscr{P}^-(K)| \leq \tau \, |\mathscr{P}^-(\varnothing)| 
\end{aligned}
\end{equation}
where $|\mathscr{P}^-(\varnothing)|$ is the number of source-to-target paths in the original Lexis-DAG,
without removing any nodes.\footnote{It is easy to see that $|\mathscr{P}^-(\varnothing)|$ is equal to the cumulative 
length of all target strings $L$.}

Note that if $\tau=0$ the core identification problem becomes equivalent to finding the min-vertex-cut of 
the given Lexis-DAG. 
In practice a Lexis-DAG often includes some {\em tendril-like} source-to-target paths traversing
a small number of intermediate nodes that very few other paths traverse. These paths can cause a large increase in
the size of the core. For this reason, we prefer to consider the case of a positive, but potentially small, 
value of the threshold $\tau$.   

%
We solve the core identification problem with a greedy algorithm referred to as \textsc{G-Core}. 
This algorithm adds in each iteration the node with the highest path-centrality value to the core set, 
updates the Lexis-DAG by removing that node and its edges, and recomputes the path centralities before
the next iteration. The algorithm terminates when the desired fraction of source-to-target paths has been 
achieved.
\textsc{G-Core} requires at most $O(|V|)$ iterations, and in each iteration we update the 
path centralities in $O(|E|)$ time.  So the run-time complexity of \textsc{G-Core} is $O(|V||E|)$.

\vspace{-2mm}
\section{Applications of Lexis}
We now discuss a variety of applications of the proposed framework. Note that in all experiments, we use the library from \cite{galle} for extracting the maximal repeats and NetworkX \cite{networkx} as the graph library in our implementation.
\subsection{Optimized String Hierarchies}
Lexis can be used as an optimization tool for the hierarchical synthesis of sequences.
One such application comes from synthetic biology, where novel DNA sequences 
are created by concatenating existing DNA sequences in a hierarchical process \cite{dnaSynthPaperNew}.
The cost of DNA synthesis is considerable today due to the biochemical operations that
are required to perform this ``genetic merging'' \cite{dnaSynthPaperNew,dnaSynthPaper}. 
Hence, it is desirable to re-use existing DNA sequences, and more generally, 
to perform this design process in an efficient hierarchical manner. 

Biologists have created a library of synthetic DNA sequences, referred to as iGEM \cite{igem}. 
Currently, there are $787$ elementary ``BioBrick parts'' from which longer composite sequences can be created.
Longer sequences are submitted to the Registry of Standard Biological Parts in the annual iGEM competition, then functionally evaluated and labeled.
In the following, we analyze a subset of the iGEM dataset. 
In particular, this dataset contains $1,375$ composite DNA sequences that are labeled as
iGEM {\em devices} because they have distinct biological functions; we treat these sequences as Lexis targets.  
The cumulative length of the target sequences is $6,957$ symbols. 
The $787$ elementary BioBrick parts are treated as the Lexis sources.
The iGEM dataset also includes other BioBrick parts that are neither devices nor elementary, and that have been used to construct more complex parts in iGEM; we ignore those because they do not have a distinct biological function (i.e., they should not be viewed as targets but as intermediate sequences that different teams of biologists have previously constructed).

We constructed an optimized Lexis-DAG for the given sets of iGEM sources and targets. 
To quantify the gain that results from using a hierarchical synthesis process, we 
compare the number of edges and concatenations in the Lexis-DAG versus a flat synthesis
process in which each target is independently constructed from the required sources. 
The {\em Lexis} solution requires only 52\% of the edges (or 56\% of the concatenations) that the flat process would require. 
The sequence with the highest path centrality in the Lexis-DAG is \emph{B0010-B0012}.\footnote{
BioBricks start with \emph{BBa\_} prefix that are omitted here.}
This part is registered as \emph{B0015} in the iGEM library and it is the most common ``terminator'' in
iGEM devices. {\em Lexis} identified several more high centrality parts that are already in iGEM, 
such as \emph{B0032-E0040-B0010-B0012}, registered as \emph{E0240}.
Interestingly, however, the Lexis-DAG also includes some high centrality parts that have not been registered
in iGEM yet, such as \emph{B0034-C0062-B0010-B0012-R0062}.
A list of the top-15 nodes in terms of path centrality is given in the Appendix.

To explore the hierarchical nature of the iGEM sequences, we compared the {\em ``Original''} Lexis-DAG,
the one we constructed with the actual iGEM devices as targets, with {\em ``Randomized''} Lexis-DAGs.
{\em ``Randomized''} Lexis-DAG is the result of applying \textsc{G-Lexis} to a target set where each iGEM device sequence is randomly reshuffled.
We compare the Original Lexis-DAG characteristics to the average Lexis-DAG characteristics over ten randomized experiments. 
The Original Lexis-DAG has fewer intermediate nodes than the Randomized ones 
(169 in Original vs 359 in Randomized), and its depth is twice as large (8 vs 4.4).
Importantly, the Randomized DAGs are significantly more costly: 44\% higher cost in terms of edges
and 52\% in terms of concatenations.  
  
\begin{figure}[t]
\centering
\begin{subfigure}[b]{.235\textwidth}
\includegraphics[width=\textwidth, trim = 1cm 6cm 3cm 6cm, clip]{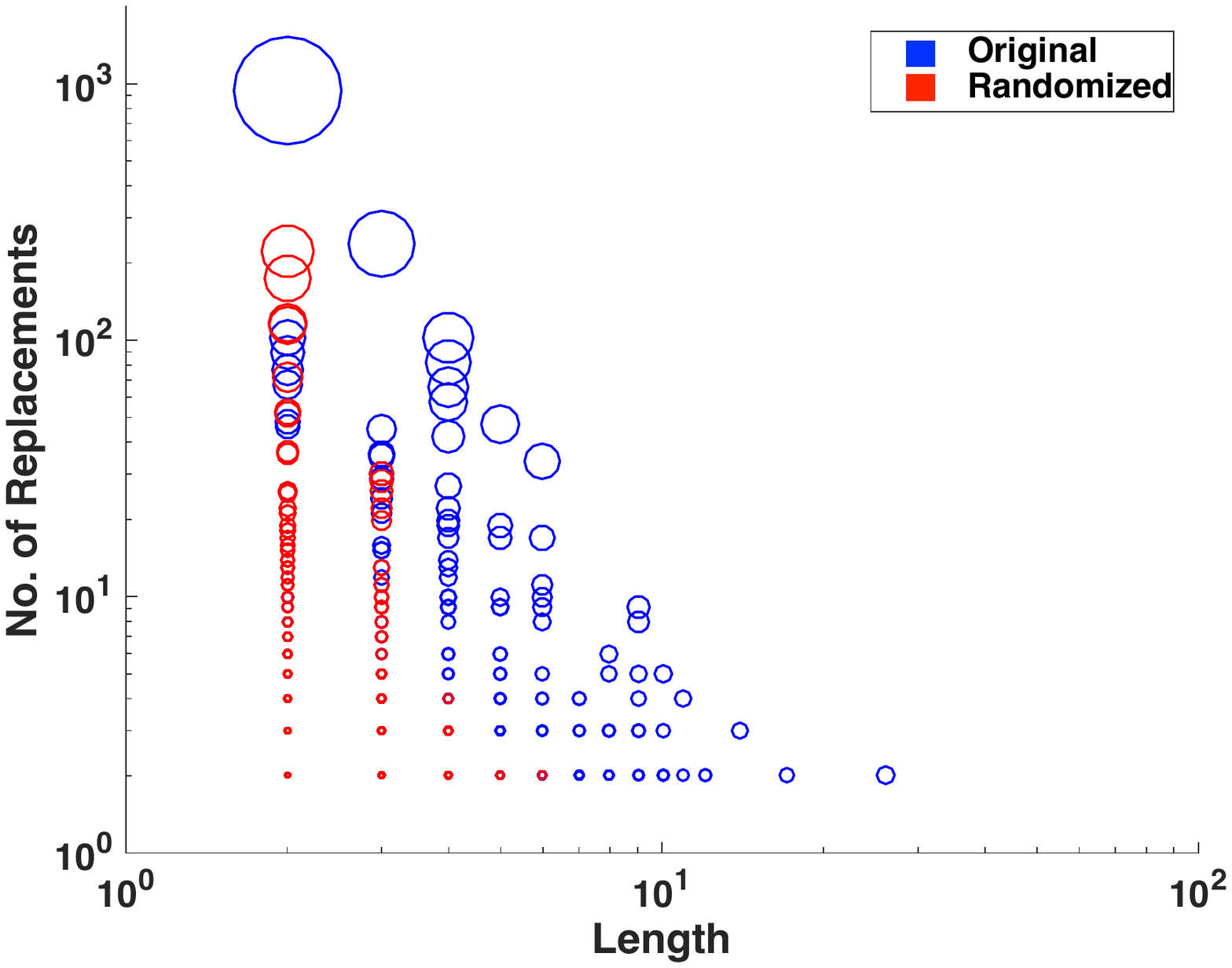}
\end{subfigure}
\begin{subfigure}[b]{.235\textwidth}
\includegraphics[width=\textwidth, trim = 1cm 6cm 3cm 6cm, clip]{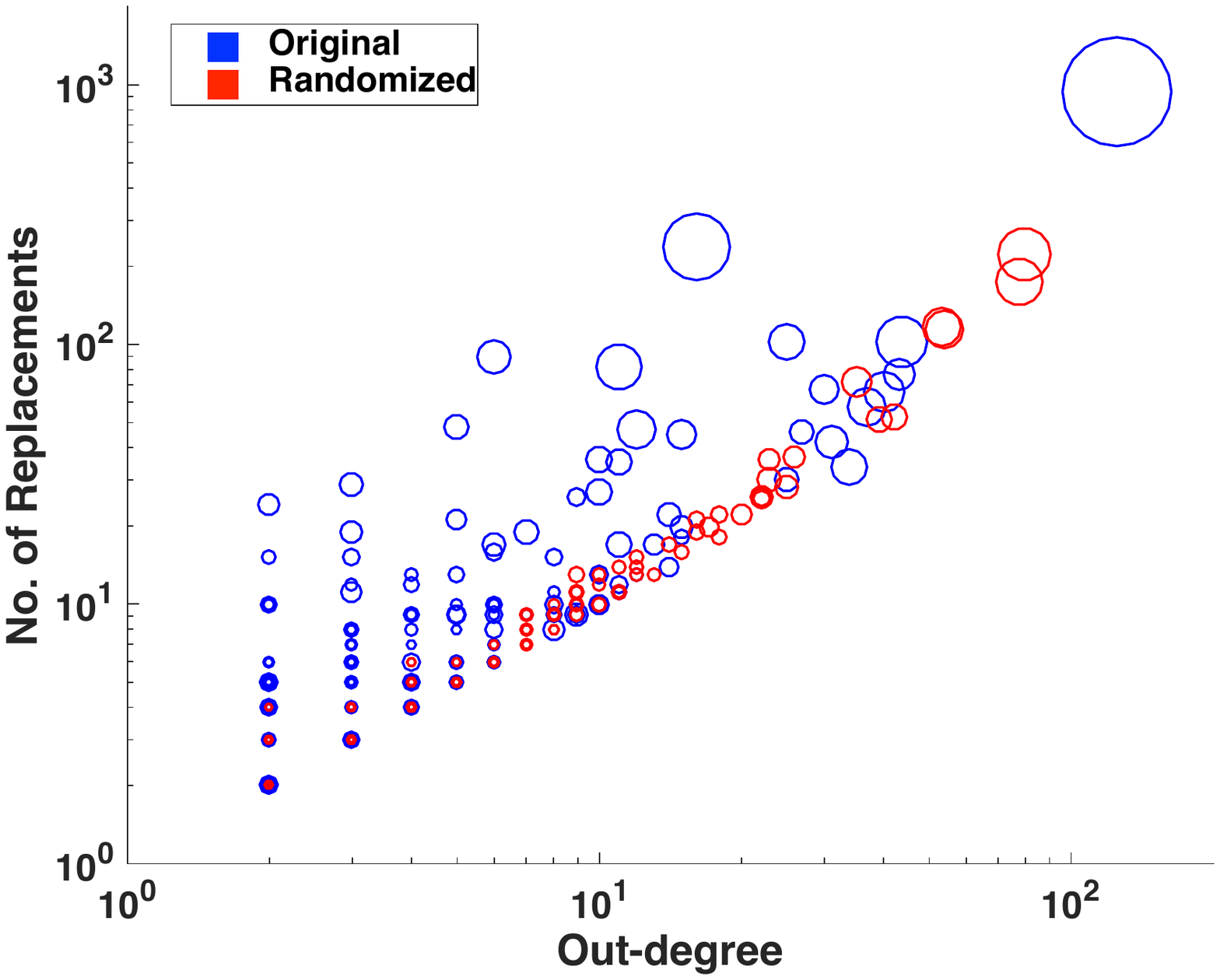}
\end{subfigure}
\caption{\small Comparison of the Lexis-DAGs that result from the Original and Randomized iGEM devices (best viewed in color).
The size of each point represents the path-centrality of that node.
}
\label{lenoutpc}
\vspace{-4mm}
\end{figure}

To further understand these differences from the topological perspective,  
Fig.~\ref{lenoutpc} shows scatter plots for the length, path centrality, and re-use (number of replacements)
of each intermediate node in the Original Lexis-DAG vs one of the Randomized Lexis-DAGs.
With randomized targets, the intermediate nodes are short (mostly 2-3 symbols), their re-use is 
roughly equal to their out-degree, and their path centrality is determined by their out-degree; 
in other words, most intermediate nodes are directly connected to the targets that include them, 
and the most central nodes are those that have the highest number of such edges.
On the contrary, with the original targets we find longer intermediate nodes (up to 11-12 symbols) 
and their number of replacements in the targets can be up to an order of magnitude higher than their out-degree.
This happens when intermediate nodes with a large number of replacements are not only used directly to 
construct targets
but they are repeatedly combined to construct longer intermediate nodes, creating a deeper hierarchy of re-use.  
In this case, the high path centrality nodes tend to be those that are both relatively long and common,
achieving a good trade-off between specificity and generality.

\subsection{Structure Discovery}
As mentioned in the introduction, it is often the case that the hierarchical process that creates 
the observed sequences is unknown. 
{\em Lexis} can be used to discover underlying hierarchical structure
as long as we have reasons to believe that that hierarchical process cares to minimize, 
even heuristically, the same cost function that Lexis considers (i.e., number of edges or concatenations). 
A related reason to apply Lexis in the analysis of sequential data is to identify the 
most parsimonious way, in terms of number of edges or concatenations, to represent the
given sequences hierarchically. Even though this representation may not be related to the process
that generated the given targets, it can expose if the given data have an inherent hierarchical structure. 

As an illustration of this process, we apply Lexis on a set of protein sequences. 
Even though it is well-known that such sequences include conserved and repeated subsequences (such as motifs) of various
lengths, it is not currently known whether these repeats form a hierarchical structure. 
That would be the case if one or more short conserved sequences 
are often combined to form longer conserved sequences, which can 
themselves be combined with others to form even longer sequences, etc. 
If we accept the premise that a conserved sequence serves a distinct biological function, 
the discovery of hierarchical structure in protein sequences would suggest that elementary 
biological functions are combined in a Lego-like manner to construct the complexity and diversity of the proteome. 
In other words, the presence of hierarchical structure would suggest that proteins satisfy, 
at least to a certain extent, the {\em composability} principle,
meaning that the function of each protein is composed of, and it can be understood through, 
the simpler functions of hierarchical components.

Our dataset is the proteome of baker's Yeast\footnote{\url{http://www.uniprot.org/proteomes/UP000002311}}, which
consists of 6,721 proteins. 
However, this includes many protein homologues.
It is important that we replace each cluster of homologues with a single protein; otherwise Lexis can 
detect repeated sequences within two or more homologues.
To remedy this issue, we use the {\em UCLUST} sequence clustering tool \cite{uclust}, 
which is based on the {\em USEARCH} similarity measure (or identity search) \cite{usearch}.
The Percentage of Identity (PID) parameter controls how similar two sequences should be so that they are 
assigned to the same cluster.
We set PID to 50\%, which reduces the number of proteins to 6,033. 
Much higher PID values do not cluster together some obvious homologues, while lower PID values are too restrictive.\footnote{\url{http://drive5.com/usearch/manual/uclust_algo.html}}
To reduce the running time associated with the randomization experiments described next,
we randomly sample 1,500 proteins from the output of {\em UCLUST}.

\begin{figure}[h]
\centering
\includegraphics[width=0.23\textwidth, trim = 1.8cm 6.5cm 2.4cm 8cm, clip]{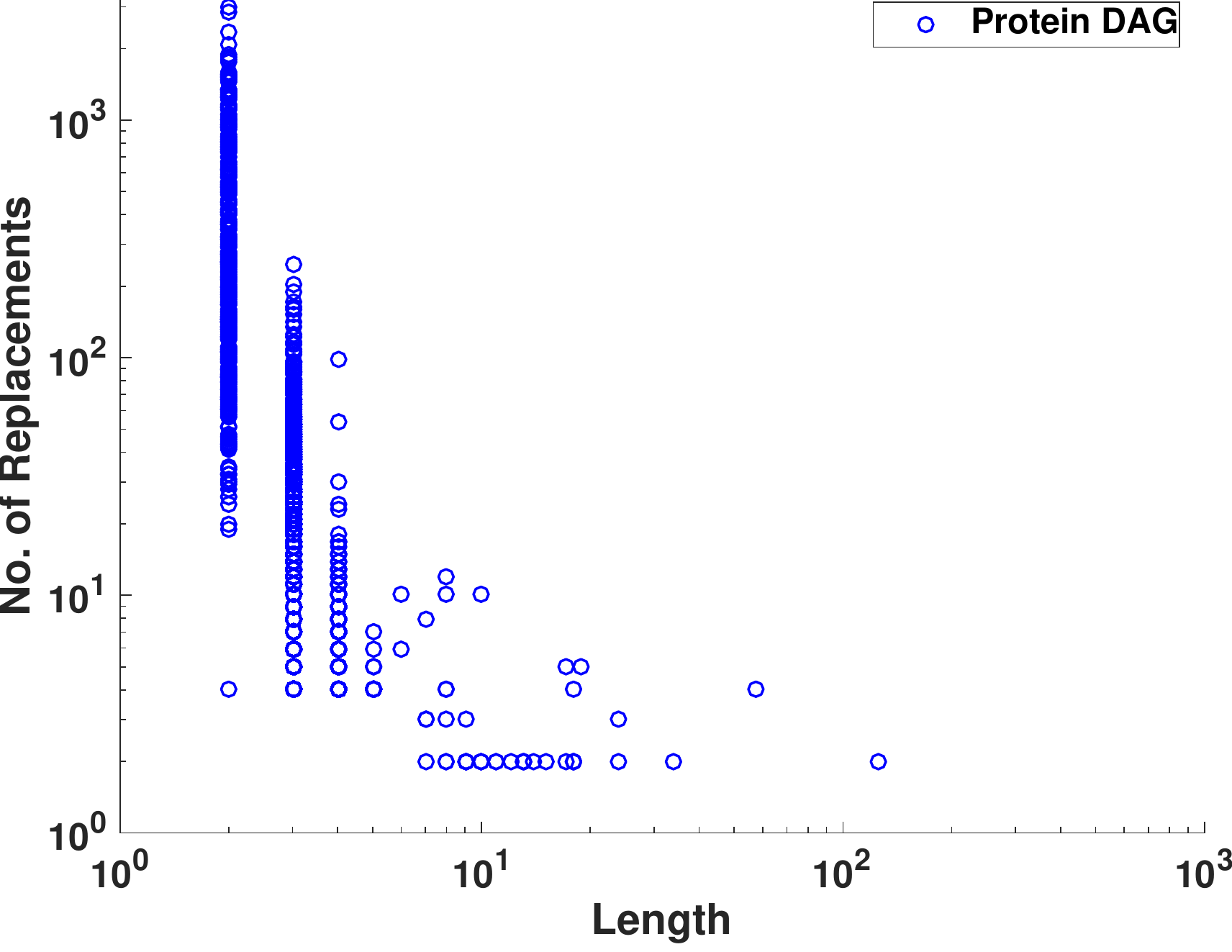}
~~\includegraphics[width=0.23\textwidth, trim = 1.8cm 6.5cm 2.4cm 8cm, clip]{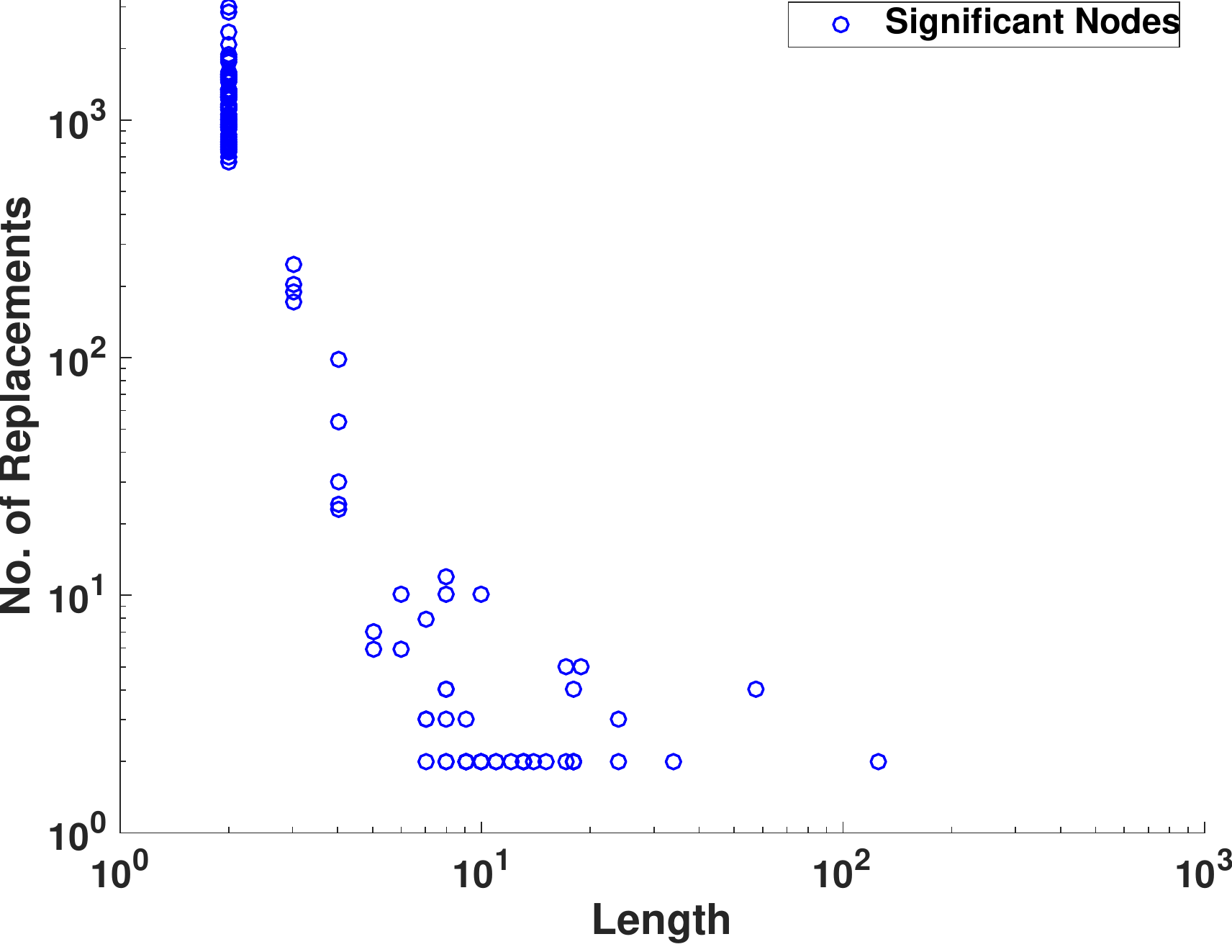}
\\
(a)~~~~~~~~~~~~~~~~~~~~~~~~~~~~~~~~~~~~~(b)
\caption{\small Length and number of replacements for the intermediate nodes in the protein sequence Lexis-DAG (a) before and (b) after we filter out the nodes that are not statistically significant. 
}
\label{yeastScatter}
\end{figure}

The total length of the protein targets is about 344K amino acids. 
The resulting Lexis-DAG has about 151K edges and 5,171 intermediate nodes, and its maximum depth is 7. 
Fig. \ref{yeastScatter}(a) shows a scatter plot of the length and number of replacements of these intermediate Lexis
nodes (repeated sequences discovered by Lexis). 

Of course some of these sequences may not have any biological significance because their length and number of 
replacements may be so low that they are likely to occur just based on chance.
For instance, a sequence of two amino acids that is repeated just twice in a sequence of thousands of amino acids
is almost certain (the distribution of amino acids is not very skewed).
To filter out the sequences that are {\em not} statistically significant, we rely on the following hypothesis test.
Consider a node that corresponds to a sequence with length $l$ and number of replacements $r$ 
in the given targets.
The null-hypothesis is that sequences with these values of $l$ and $r$ will occur in a Lexis-DAG that is constructed 
for a random permutation of the given targets.
To evaluate this hypothesis,
we randomize the given target sequences multiple times, and construct a Lexis-DAG for each randomized sequence. 
We then estimate the probability that sequences of length $l$ and number of replacements $r$ occur in the
randomized target Lexis-DAG, as the fraction of 500 experiments in which this is true.
For a given significance level $\alpha=0.1$, we can then identify the pairs $(l,r)$ for which we can
reject the null-hypothesis; these pairs correspond to the nodes that we view as statistically
significant.\footnote{Another way to conduct this hypothesis test would be to estimate the probability that 
a specific sequence of length $l$ will be repeated $r$ times in a permutation of the targets.
The number of randomization experiments would need to be much higher in that case, however, to cover all
sequences that we see in the actual Lexis-DAG, each with a given value of $r$.} 

On average, the randomized target Lexis-DAGs have a smaller depth ($5.0$) and more edges ($155K$) than the original Lexis-DAG.
Fig. \ref{yeastScatter}(b) shows the intermediate nodes of the original Lexis-DAG that 
pass the previous hypothesis test.

\begin{figure*}[t]
\centering
\includegraphics[width=\textwidth, trim = 1cm 7.2cm 1cm 7.2cm, clip]{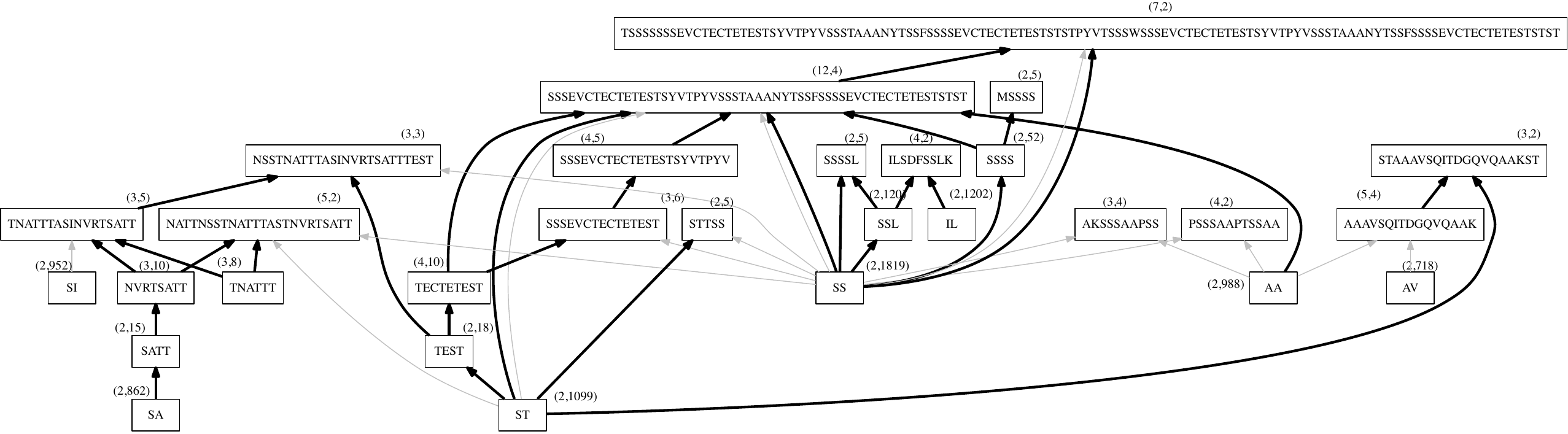}
\caption{\small A small subgraph of the Lexis-DAG for 1,500 Yeast proteins. This plot only shows statistically significant nodes. Bold edges represent a direct connection between two nodes, while grey edges represent indirect connections (through nodes that are not shown in this plot). The label on top of each node shows the tuple (in-degree, number of replacements).} 
\label{yeastGraph}
\vspace{-3.5mm}
\end{figure*}

Fig. \ref{yeastGraph} shows a small subgraph of the Lexis-DAG, showing only about 30 intermediate nodes; all
these nodes have passed the previous significance test. 
The grey lines represent indirect connections, going through nodes that have not passed the significance test
(not shown), while the bold lines represent direct connections. 
Interestingly, there seems to be a non-trivial hierarchical structure with several long paths, 
and with sequences of several amino acids that repeat several times even in this relatively small sample of proteins. 
Despite these preliminary results, it is clearly 
still early to draw any hard conclusions about the presence of hierarchical structure in protein sequences.
We are planning to further pursue this question using Lexis in collaboration with domain experts. 
  

%
%


\subsection{Compression}
Recent work has highlighted the connection between pattern mining and 
sequence compressibility \cite{gokrimp}. 
Data compression looks for regularities that can be used to compress the data, 
while patterns are often useful as such regularities. 
In dictionary-based lossless compression, the original sequence is encoded with the help of a dictionary, 
which is a subset of the sequence's substrings 
as well as a series of pointers indicating the location(s) at which each dictionary element 
should be placed to fully reconstruct the original sequence. 
Following the Minimum Description Principle, 
one strives for a compression scheme that results in the smallest size for the joint representation of both
the dictionary and the encoding of the data using that dictionary. 
The size of this joint representation is the total space needed to store the
dictionary entries plus the total space needed for the required pointers.
We assume for simplicity that the space to store an individual character and a pointer are the same.

We now evaluate the use of a Lexis-DAG for compression (or compact representation) of strings. 
To do so, we need to decide 1) how to choose the patterns that will be used for compression, 
and 2) if a pattern appears more than once, which occurrences of that pattern to replace. 
A naive approach is to simply use the set of substrings that appear in the Lexis-DAG 
as dictionary entries, and compare them to sets of patterns found by other substring mining algorithms.
Given a set of patterns as potential dictionary entries, 
selecting the best dictionary and pointer placement is NP-hard. 
A simple greedy compression scheme, that we refer to as {\em CompressLR}, 
is to iteratively add to the dictionary the substring that gives the highest compression gain 
when replacing all Left-to-Right non-overlapping occurrences of that substring with pointers. 
We re-evaluate the compression gain of candidate dictionary entries in each iteration.
For a substring $v$ with length $|v|$ and number of left-to-right non-overlapping occurrences $R_v$, 
the compression gain is: 
\begin{equation}
R_v \times |v| - R_v - |v| = \left(R_v-1\right) \times \left(|v| - 1\right) - 1
\end{equation}

We compare the substrings identified by Lexis with the substrings generated by 
a recent contiguous pattern mining algorithm called 
{\em ConSgen} \cite{consgen} (we could only run it on the smallest available dataset).
Additionally, we compare the {\em Lexis} substrings with the set of patterns containing all 2- and 3-grams of the data. 
The comparisons are performed on six sequence datasets: the ``Yeast'' and iGEM datasets of the 
previous sections, as well as four ``NSF CS awards'' datasets that will be described in more detail
in the next section.  

Table \ref{tab:compress} shows the comparison results under the 
headings:  Lexis-CompressLR, 2+3grams-CompressLR and ConSgen-CompressLR. 
These naive approaches are all on par with each other.
This comparison, however, treats \textsc{G-Lexis} as a mere pattern mining algorithm. 
Instead, the \textsc{G-Lexis} algorithm constructs a Lexis-DAG that puts the generated patterns in a hierarchical context. 
One can think of the Lexis-DAG as the instructions in constructing a hierarchical ``Lego-like'' sequence. 
The edges into the targets tell us how to place the final pointers, i.e., which occurrences of a dictionary entry to replace
in the targets. 
Further, the rest of the DAG shows how to compress the patterns that appear in the targets using smaller patterns. 
It is easy to see that using this strategy the compressed size becomes equal to the number of edges in the DAG. 
Using this strategy that is encoded in the Lexis-DAG results in an additional 2\%-20\% reduction 
in the compressed size over the CompressLR approaches.

\begin{table}[h]
{\scriptsize
\begin{tabular}{c"c|c c c}
\multirow{3}{*}{\bf Dataset} & \multirow{3}{*}{{\bf Lexis}} &\multirow{3}{*}{\bf Lexis}&\multirow{3}{*}{\bf 2+3-gram}&\multirow{3}{*}{\bf ConSgen}
\\&&&{\bf}
\\
{\bf }&{\bf DAG}&{\bf CompressLR}&{\bf CompressLR}&{\bf CompressLR}
\\\thickhline
\multirow{2}{*}{\tiny \bf Know\&Cog}  & \multirow{2}{*}{\bf 68.69} & \multirow{2}{*}{76.58} & \multirow{2}{*}{77.13} & \multirow{2}{*}{---}
\\
\multirow{2}{*}{\tiny \bf Networks} & \multirow{2}{*}{\bf 78.48} & \multirow{2}{*}{86.47}& \multirow{2}{*}{86.43}& \multirow{2}{*}{---}
\\
\multirow{2}{*}{\tiny \bf Robotics} & \multirow{2}{*}{\bf 73.19} & \multirow{2}{*}{80.62}& \multirow{2}{*}{79.69}& \multirow{2}{*}{---}
\\
\multirow{2}{*}{\tiny \bf Theory} & \multirow{2}{*}{\bf 79.41} & \multirow{2}{*}{81.89}& \multirow{2}{*}{82.63}& \multirow{2}{*}{---}
\\
\multirow{2}{*}{\tiny \bf Yeast}  & \multirow{2}{*}{\bf 44.28} & \multirow{2}{*}{51.08} &\multirow{2}{*}{50.71} &\multirow{2}{*}{---}
\\
\multirow{2}{*}{\tiny \bf iGEM}  & \multirow{2}{*}{\bf 47.86} & \multirow{2}{*}{67.47} &\multirow{2}{*}{67.75} &\multirow{2}{*}{67.47}
\\
&&&
\\
\end{tabular}}
\footnotesize
\centering
\vspace{-3mm}
\caption{\small Compression ratio (i.e., percentage of compressed data size over original data size).} 
\label{tab:compress}
\end{table}

\subsection{Feature Extraction}
The Lexis-DAG can also be used to extract machine learning features for sequential data. 
The intermediate nodes that form the core of a Lexis-DAG, in particular,  correspond 
to sequences that are both relatively long and frequently re-used in the targets. 
We hypothesize that such sequences will be good features for machine learning tasks
such as classification or clustering because they can discriminate different 
classes of objects (due to their longer length) and at the same time they   
are general within the same class of objects (due to their frequent occurrence in the targets of that class).
 
To test this hypothesis, we used Lexis to extract text features for four classes of NSF research award
abstracts during the 1990--2003 time 
period.\footnote{\url{archive.ics.uci.edu/ml/machine-learning-databases/nsfabs-mld/nsfawards.data.html}}
We pre-processed each award's abstract through stopword removal and Porter stemming. 
The alphabet $S$ is the set of word stems that appear at least once in any of these abstracts. 
Table~\ref{nsfData} describes this dataset in terms of number of abstracts, cumulative abstract length,
and average length per abstract for each class.

\begin{table}[h]
\scriptsize
\centering
\caption{\small Description of 4 classes of NSF award abstracts}
\vspace{-3mm}
\begin{tabular}{c"c c P{1.2cm} P{1.54cm}}
\multirow{3}{*}{{\bf Class}} & \multirow{3}{*}{{\bf $|T|$}} & \multirow{3}{*}{{\bf $L$}} & \multirow{3}{*}{{\bf $L/|T|$}} & \multirow{3}{*}{{\bf $|V_M|$ }}
\\
&&&
\\\thickhline
\multirow{3}{*}{{\bf Knowledge}} & \multirow{3}{*}{411} & \multirow{3}{*}{47,858} & \multirow{3}{*}{116}& \multirow{3}{*}{2,902}
\\
&&&
\\
{\bf \&Cog Sci} & &
\\
{\bf Networks} & 836 & 74,738 & 89 & 3,730
\\
{\bf Robotics} & 496 & 56,481 & 113 & 4,560
\\
{\bf Theory} & 566 & 66,891 & 118 & 4,247
\end{tabular}
\label{nsfData}
\vspace{-3mm}
\end{table}

We constructed the Lexis-DAG for each class of abstracts, 
and then used the \textsc{G-Core} algorithm to identify the core for each DAG. We stopped \textsc{G-Core} at the point where $95\%$ of indirect paths in the Lexis-DAG are covered.
The strings in each core are the extracted features for the corresponding class of abstracts. 
Table~\ref{topFeat} shows the 5 strings extracted by \textsc{G-Core} for each class. 
We create a common set of \textsc{G-Core} features by taking the union of the sets of core substrings derived for each class. 
The next step is to construct the feature vector for each abstract. 
We do so by representing each abstract as a vector of counts, with a count for each substring feature.

\begin{table}[h]
\vspace{-2mm}
\footnotesize
\centering
\caption{\small 5 word stems in the Lexis-DAG core of each class of abstracts.}
\vspace{-2.5mm}
\begin{tabular}{c"c}
{\bf Knowledge \& Cog} & {\bf Networks}
\\\thickhline
machine learn & request support nsf connect
\\
knowledg base & bit per second
\\
natur languag & two year
\\
artifici intellig & provide partial support
\\
neural network & high perform
\\
&
\\
{\bf Robotics} & {\bf Theory}
\\\thickhline
comput vision & comput scienc
\\
first year & real world
\\
robot system & complex class
\\
object recognit & complex theori
\\
real time & approxim algorithm
\end{tabular}
\label{topFeat}
\end{table}

\begin{table}[h]
\vspace{-5mm}
\footnotesize
\centering
\caption{SVM classification accuracy. \# nonzeros is the number of nonzero elements in the term-document matrix with each feature set. The accuracies and the parameters are fit based on 10-fold cross-validation for each feature set.} 
{\scriptsize
\begin{tabular}{l"c|c|c}
\multirow{1}{*}{\bf Method (\# Features)} & \multirow{1}{*}{{\bf \# Nonzeros }} & {\bf ($\gamma,c$)} & {\bf Accuracy}
\\\thickhline
\multirow{2}{*}{\tiny \bf Bag-of-words (9,7k)} & \multirow{2}{*}{190,2k}  & \multirow{2}{*}{\emph{(0.0015,3)}} & \multirow{2}{*}{88.3\%}
\\
\multirow{2}{*}{\tiny \bf \textsc{G-Core} (14,4k)}  & \multirow{2}{*}{{\bf 55,5k}} & \multirow{2}{*}{\emph{(0.02,1)}} & \multirow{2}{*}{90.0\%}
\\
\multirow{2}{*}{\tiny \bf 2-Gram (124,9k)} & \multirow{2}{*}{228,0k}  & \multirow{2}{*}{\emph{(0.0015,3)}} & \multirow{2}{*}{\bf 91.3\%}
\\
\multirow{2}{*}{\tiny \bf 3-Gram (186,3k)} & \multirow{2}{*}{234,5k} & \multirow{2}{*}{\emph{(0.001,1)}} & \multirow{2}{*}{75.8\%}
\\
\multirow{2}{*}{\tiny \bf 1+2-Gram (134,6k)}  & \multirow{2}{*}{418,2k} & \multirow{2}{*}{\emph{(0.001,1)}} & \multirow{2}{*}{90.9\%}
\\
\multirow{2}{*}{\tiny \bf 1+2+3-Gram (321,0k)} & \multirow{2}{*}{652,8k} & \multirow{2}{*}{\emph{(0.001,1)}} & \multirow{2}{*}{89.2\%}
\\
\end{tabular}}
\label{svm}
\end{table}

To assess how good these features are, we compare the classification 
accuracy obtained using the Lexis features with more mainstream representations
in text mining on NSF data:  ``bag-of-words'', 2-gram, 3-gram, and two combinations of these representations. 
We use a basic SVM classifier with an RBF kernel. We used the SVM implementation in MATLAB.
The accuracy results are similar to those with a KNN classifier that we tried with a Cosine distance, 
and the accuracy is evaluated with 10-fold cross-validation.

Table~\ref{svm} shows that the Lexis features result in a much sparser term-document 
matrix, and so in smaller data overhead in learning tasks, without sacrificing classification accuracy. 
Lexis also results in a lower feature dimensionality (with the exception of the 1-gram method but the
accuracy of that method is much lower).
Note that most {\em Lexis} features are 2-grams but the {\em Lexis}  core (for $95\%$ of path coverage) may also include 
longer $n$-grams.
Lexis becomes better, relative to the other feature sets,
as we decrease the number of considered features. 
For instance, with $3,000$ features the accuracy with Lexis is 74\%, 
while the accuracy with the 1-gram and 2-gram features is 69\% and 64\%, respectively.


\section{Related Work}
Lexis is closely related to the \emph{Smallest Grammar Problem} (SGP), which focuses on the following question:
{\em What is the smallest context-free grammar that only generates a given string?} 
The constraint that the grammar should generate only one string is important because 
otherwise we could simply consider a $\Sigma^*$ as the generator of any string over $\Sigma$. 
The SGP is NP-hard, and inapproximable beyond a ratio of $\frac{8569}{8568}$ \cite{sgp}.
Algorithms for SGP have been used for string compression \cite{repair} and structure discovery \cite{sequitur} (for a survey see \cite{galle}).
There are major differences between {\em Lexis} and SGP. 
First, in {\em Lexis} we are given a set of several target strings, not only one string.
Second, {\em Lexis} infers a network representation (a DAG) instead of a grammar,
and so it is a natural tool for questions relating to the hierarchical structure of the targets. 
For instance, the centrality analysis of intermediate nodes or the core identification 
question are well understood problems in network analysis, while it is not obvious how to approach them
with a grammar-based approach. 

One can also relate {\em Lexis} to the body of work on \emph{sequence pattern mining}, 
where one is interested in discovering frequent or interesting patterns in a given sequence. 
Most work in this area has focused on mining subsequences, i.e., a set of ordered but not necessarily 
adjacent characters from the given sequence. 
In Lexis, we focus on identifying substrings, also known as {\em contiguous sequence patterns}. 
A couple of recent papers develop algorithms for mining substring patterns \cite{consgen,ccspan}, 
since sequence mining algorithms do not readily apply to the contiguous case. 
However, they rely on the methodology of candidate generation (commonly used in sequence pattern mining), 
where all patterns meeting a certain criterion are found, such as having frequency of at least two or being maximal. 
In the sequence mining literature, it has been recently observed that the size of 
the discovered set of patterns as well as their redundancy can be better controlled by mining for a set of patterns 
that meet a criterion collectively, as opposed to individually.
This is useful when these patterns are used as features in other tasks such as summarization or classification.  
Algorithms for such set-based pattern discovery have been recently developed for sequence pattern mining \cite{gokrimp,sqs}. 
In the context of substring pattern mining, Paskov et al \cite{paskov2013compressive} show how to identify a set  
of patterns with optimal lossless compression cost in an unsupervised setting, 
to be then used as features in supervised learning for classification.
In a follow-up paper \cite{dracula}, DRACULA provides a ``deep variant'' of \cite{paskov2013compressive} that is similar to Lexis, in terms of the general problem setup.
DRACULA's focus is mostly on complexity and learning aspects of the problem, while Lexis focuses on network analysis of the resulting optimized hierarchy. For instance, DRACULA considers how to take into account how dictionary strings are constructed to regularize learning problems, and how the optimal Dracula solution behaves as the cost varies.
We have shown that, although not specifically designed for feature extraction or compression, the {\em Lexis} framework
also results in a small and non-redundant set of substring patterns that can be used in classification and compression tasks.
%
 
\emph{Optimal DNA synthesis} is a new application domain, and we are only aware of the work by 
Blakes et. al.~\cite{dnaSynthPaper};
they describe DNALD, an algorithm that greedily attempts to maximize DNA re-use for multistage assembly of 
DNA libraries with shared intermediates.
Even though the {\em Lexis} framework was not specifically designed for DNA synthesis, the Lexis-DAGs can be seemlessly 
used as solutions for this task. 
In our illustrative example with the iGEM dataset, G-Lexis returns solutions with 11\% lower synthesis cost 
(equivalent to concatenation cost) than DNALD.

\emph{Structure discovery} in strings has been explored from several different perspectives.  
For example, the grammar-based algorithm SEQUITIR \cite{sequitur} presents interesting possible applications 
in natural language and musical structure identification. 
In an information-theoretic context,
Lancot et. al.~\cite{lanctot} shows how to distinguish between coding and non-coding regions by 
analyzing the hierarchical structure of genomic sequences. 
\section{Conclusions and Future Work}
Lexis is a novel optimization-based framework for exploring the hierarchical nature of sequence data.  
In this paper, we stated the corresponding optimization problems in the admittedly 
limited context of two simple cost functions (number of edges and concatenations), 
proved their NP-hardness, and proposed a greedy algorithm for the construction of Lexis-DAGs.
This research can be extended in the direction of more general cost 
formulations and more efficient algorithms.  
Additionally, we are working on an incremental version of Lexis in which 
new targets are added to an existing Lexis-DAG, 
without re-designing the hierarchy.  

We also applied network analysis in Lexis-DAGs, proposing a new centrality metric 
that can be used to identify the most important intermediate nodes, 
corresponding to substrings that are both relatively long and frequently occurring in the target sequences. 
This network analysis connection raises an interesting question: are there
certain topological properties that are common across Lexis-DAGs that have resulted from long-running
evolutionary processes? We have some evidence that one such topological property is that
these DAGs exhibit the {\em hourglass effect} \cite{akhshabi2013evolution}.

Finally, we gave four examples of how Lexis can be used in practice, 
applied in optimized hierarchical synthesis, structure discovery, compression and feature extraction.
In future work, we plan to apply Lexis in various domain-specific problems.
In biology, in particular, we can use Lexis in comparing the hierarchical structure of 
protein sequences between healthy and cancerous tissues. 
Another related application is {\em generalized phylogeny inference}, 
considering that horizontal gene transfers (which are common in viruses and bacteria)
result in DAG-like phylogenies (as opposed to trees). 
\section{Acknowledgements}
This research was supported by the
National Science Foundation under Grant No. 1319549.
\bibliographystyle{plain}
\bibliography{BIB}
\vspace{-3mm}
\section*{Appendix}

\subsection*{NP-hardness of Lexis Problem}
We prove that the Lexis problem is NP-hard through a reduction 
from the \emph{Smallest Grammar Problem} (SGP) \cite{sgp}. 

Formally, \emph{The Smallest Grammar Problem} for a string $s$ is to identify a 
\emph{Straight-Line Grammar}
(SLG)
$G^*$ such that 
$L(G^*) = \{s\}$ and $|G^*| \leq |G|$ for any other $G$ with $L(G) = \{s\}$, 
where $|G|$ denotes the size of grammar $G$. 
Charikar et al. \cite{sgp} define the size of a grammar as the cumulative length of the 
right-hand side of all rules, i.e.,  $|G|=\sum_{T\rightarrow \alpha \in \Delta}|\alpha|$
where $|\alpha|$ is the number of symbols appearing in the term $\alpha$ of a grammar rule.
Under this grammar size, Charikar et al. show that the Smallest Grammar Problem is NP-hard \cite{sgp}.

\textsc{Theorem 1.}
\emph{The Lexis problem in Eq. \eqref{Lexis} is NP-hard for $C(D)$ being edge cost (Eq. \eqref{firstDAGCost}) or concatenation cost (Eq. \eqref{secondDAGCost}).}

\emph{Proof:} 
Let us first start with edge cost.
Consider an instance of SGP in which we are given string $s$ and we are asked to 
compute an SLG $G$ such that $L(G)=\{s\}$ and $|G| \leq m$, 
where $|G| =\sum_{T\rightarrow \alpha \in \Delta}|\alpha|$. 
We reduce it to an instance of the Lexis problem with a single target string $T=\{s\}$,
in which we are asked to compute a Lexis-DAG $D$ with $\mathcal{E}(D) \leq m$.

Given a grammar $G=(\Sigma,\Gamma,S,\Delta)$ as a solution to the SGP problem, we construct a solution $D$ for the reduced Lexis problem. 
For each symbol in $\Sigma \cup \Gamma$, construct a node. 
For a non-terminal $T\in \Gamma$, we refer to the corresponding node also as $T$, 
and associate that node with the string $\mathcal{S}(T)$ that is produced by expanding rule 
$T$ according to grammar $G$. 
Also, for each rule $T\rightarrow \alpha$ in $G$, we scan $\alpha$ and add an edge in $D$ 
from every node that corresponds to a terminal or nonterminal in $\alpha$ to the node 
that corresponds to $T$ (along with the corresponding index). 
It is easy to see that $D$ is acyclic since $G$ is a straight-line grammar and that the number of edges in $D$ is: 
$\mathcal{E}(D) = \sum_{T\rightarrow \alpha \in \Delta}|\alpha| \leq m$.

Conversely, consider a Lexis-DAG $D=(V_S\cup V_M\cup V_T, E)$ which is a solution to the Lexis problem from our reduction above
, i.e., it has a single target string $s$, and $\mathcal{E}(D) \leq m$. 
We can construct a corresponding grammar $G$ for the SGP as follows. 
For each source node in $V_S$, construct a terminal in $\Sigma$.  
For each node $v$ in $V_T \cup V_M$, construct a nonterminal $NT(v)$ in $\Gamma$. 
For the single target node $v$ in $V_T$, designate $NT(v)$ as the start symbol $S$. 
For each node $v$ in $V_T \cup V_M$, add a rule in $\Delta$ with the 
right-hand side listing the corresponding $\Sigma\cup\Gamma$ symbols for all nodes in the sequence $I(v)$
(i.e., ordered as their respective strings appear concatenated in $\mathcal{S}(v)$). 
The constructed grammar is straight-line, since every nonterminal has one rule associated 
with it, and the grammar is also acyclic because the Lexis-DAG $D$ is acyclic. 
It is easy to see that $|G| \leq m$.

The NP-hardness proof of the Smallest Grammar Problem with grammar size defined as 
$|G| =\sum_{T\rightarrow \alpha \in \Delta}|\alpha|$ \cite{sgp} 
can be adapted for a modified grammar size definition, i.e.,  
$|G| = \sum_{T\rightarrow \alpha \in \Delta}\left(|\alpha|-1\right)=\left(\sum_{T\rightarrow \alpha \in \Delta}|\alpha|\right)-\left|\Delta\right|$. 
We can then use the same reduction from SGP to Lexis as in the case of edge cost 
to show that Lexis with concatenation cost is also NP-hard.
\hfill{$\blacksquare$}

\subsection*{Lexis-DAGs: Edge Cost vs Concatenation Cost}
\begin{figure}[h]
\center
%
%
\hspace{-5cm}
\begin{subfigure}[b]{0.1\textwidth}
(a)$~~~$\includegraphics[trim = 1.3cm 1.35cm 1.40cm 1.39cm,clip,scale=0.3]{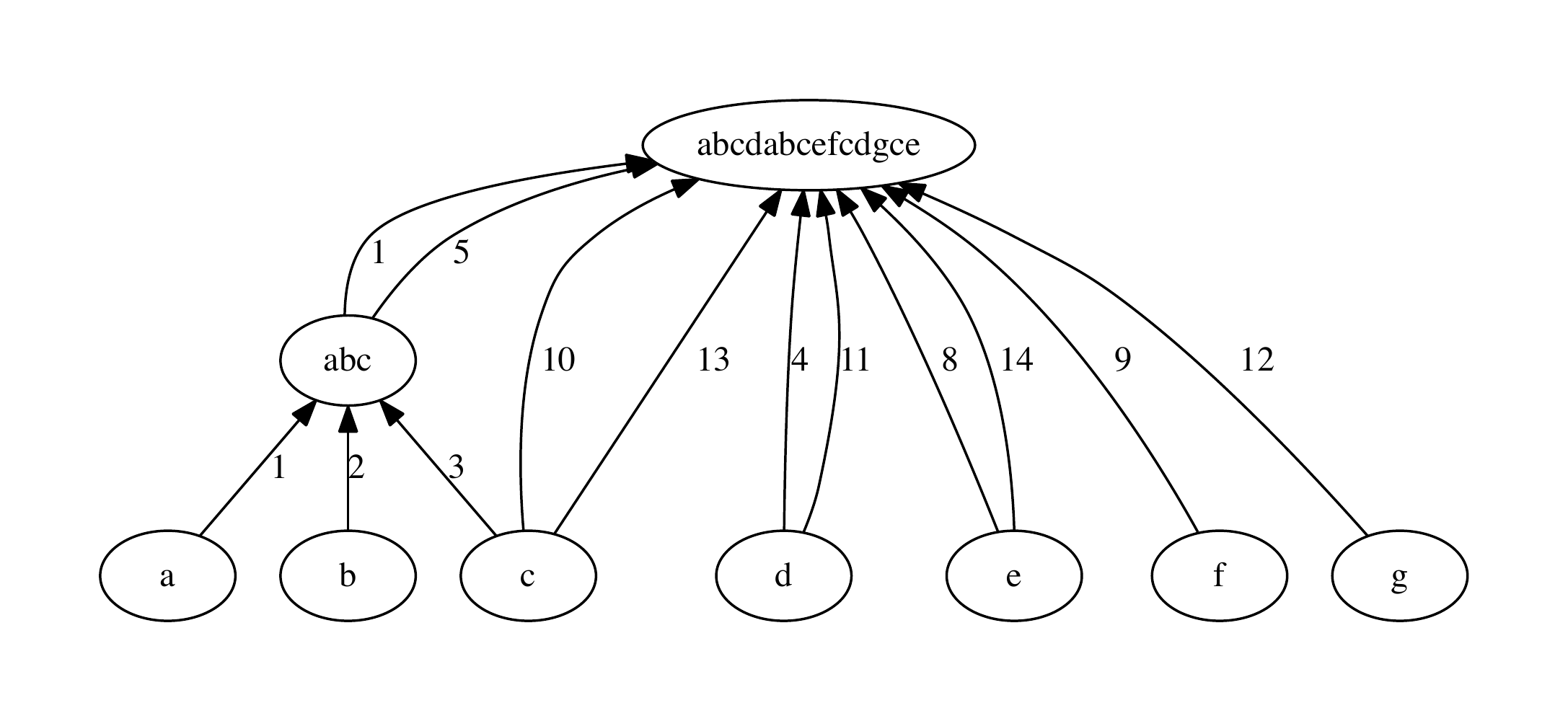}
\end{subfigure}
\\
\hspace{-2cm}\begin{subfigure}[b]{0.1\textwidth}
(b)$~~~~$\includegraphics[trim = 1.4cm 1.35cm 1.35cm 1.39cm,clip,scale=.3]{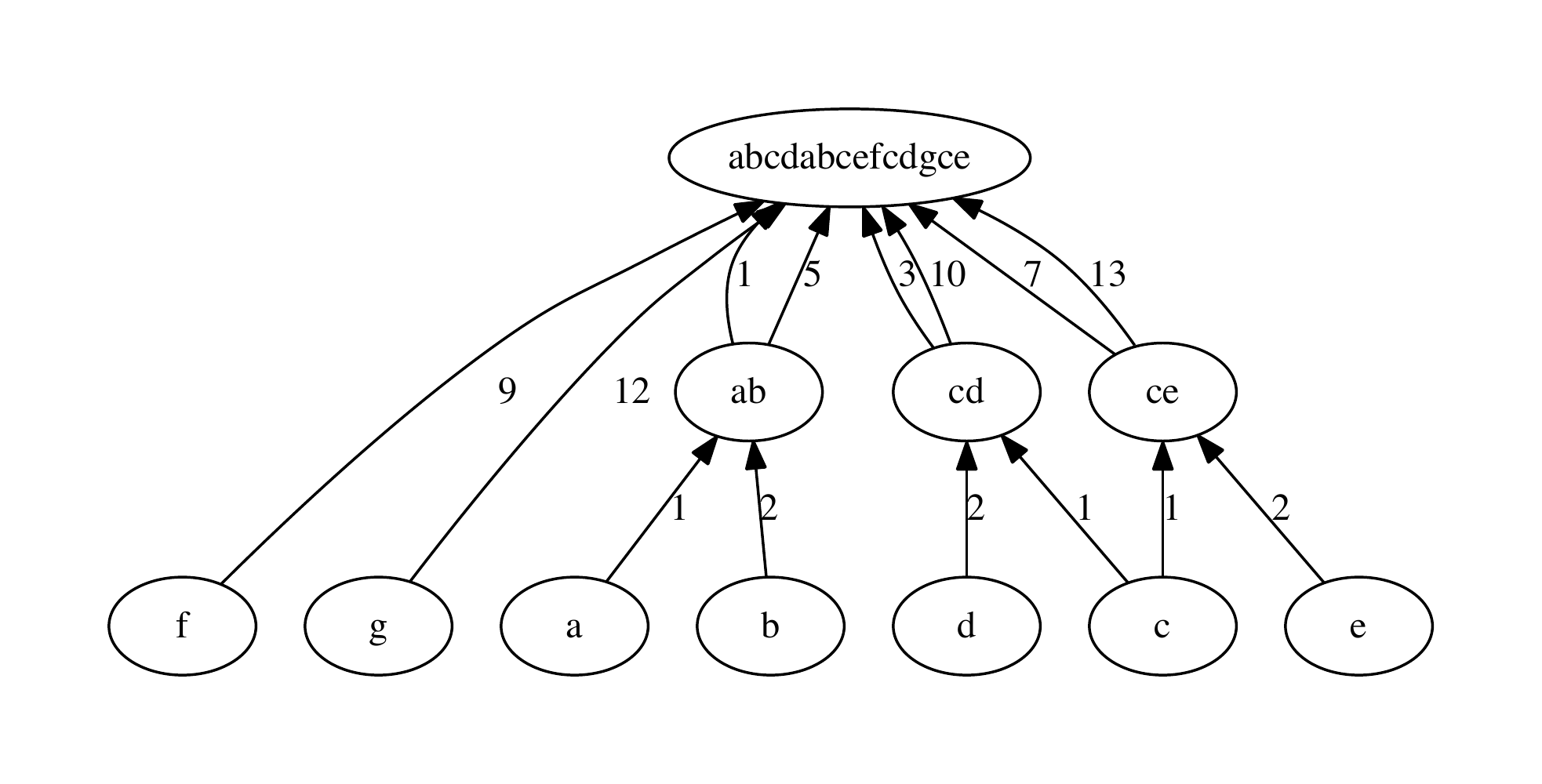}
\end{subfigure}
\caption{\footnotesize Illustration of Lexis-DAGs for target $T=\{\mathit{abcdabcefcdgce}\}$ and sources $S=\{a,b,c,d,e,f,g\}$ - {\bf (a)} Lexis-DAG $D_1$ with $\mathcal{E}(D_1)=13$ (optimal) and $\mathcal{C}(D_1)=11$ (suboptimal). {\bf (b)}  Lexis-DAG $D_2$ with $\mathcal{E}(D_2)=14$ (suboptimal) and $\mathcal{C}(D_2)=10$ (optimal).}
\label{costEx}
\end{figure}

Fig. \ref{costEx} illustrates an example that the Lexis-DAG when optimizing for 
the edge cost function may be different than the Lexis-DAG when optimizing for 
concatenation cost.

\subsection*{Comparison of \textsc{G-Lexis} with Longest Substring Replacement Algorithm}
\begin{figure}[H]
\centering
\begin{subfigure}[b]{.235\textwidth}
\includegraphics[width=\textwidth, trim = 1cm 6cm 2.5cm 6cm, clip]{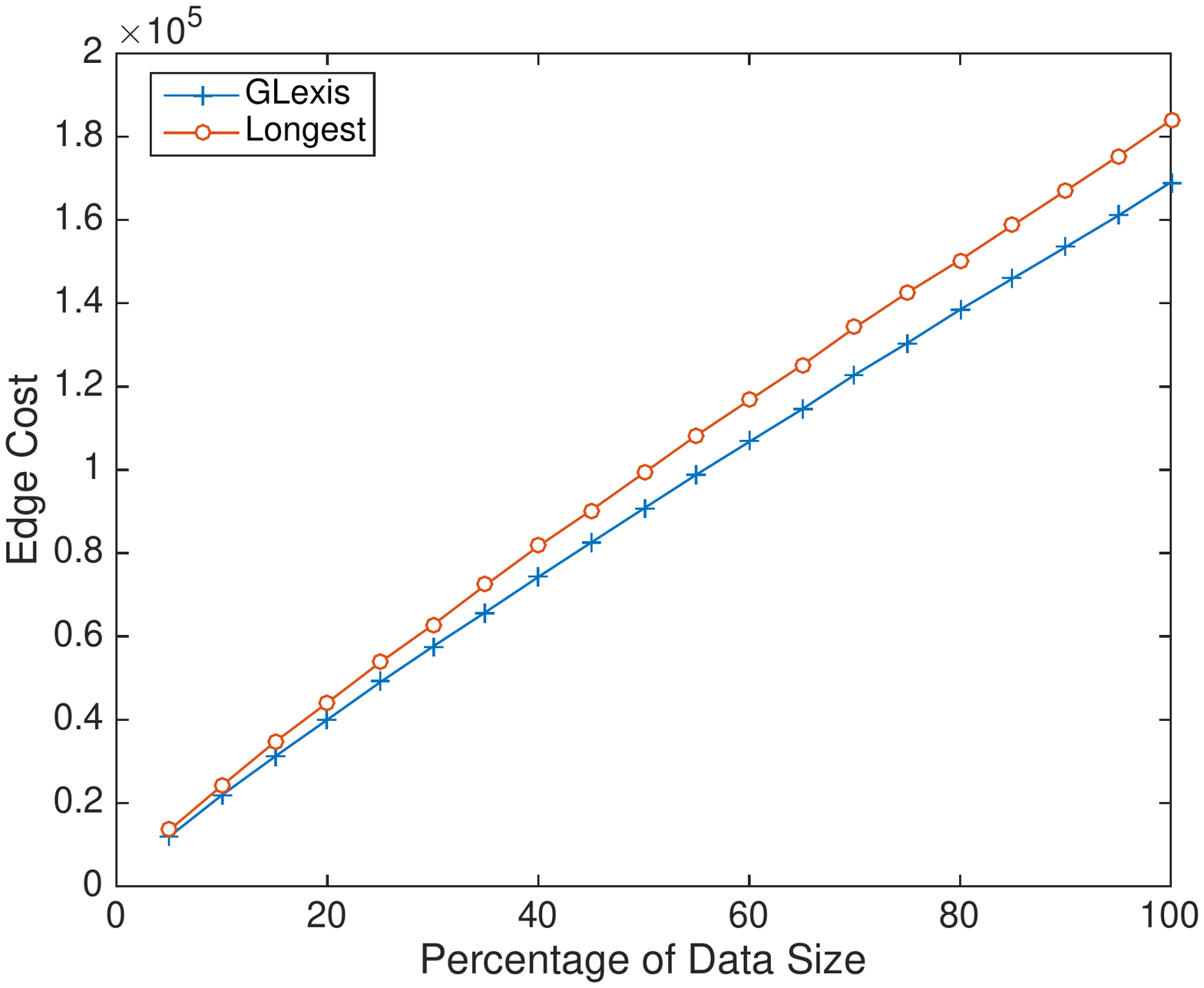}
\caption{Cost Comparison}
\end{subfigure}
\begin{subfigure}[b]{.235\textwidth}
\includegraphics[width=\textwidth, trim = 1cm 6cm 2.5cm 6cm, clip]{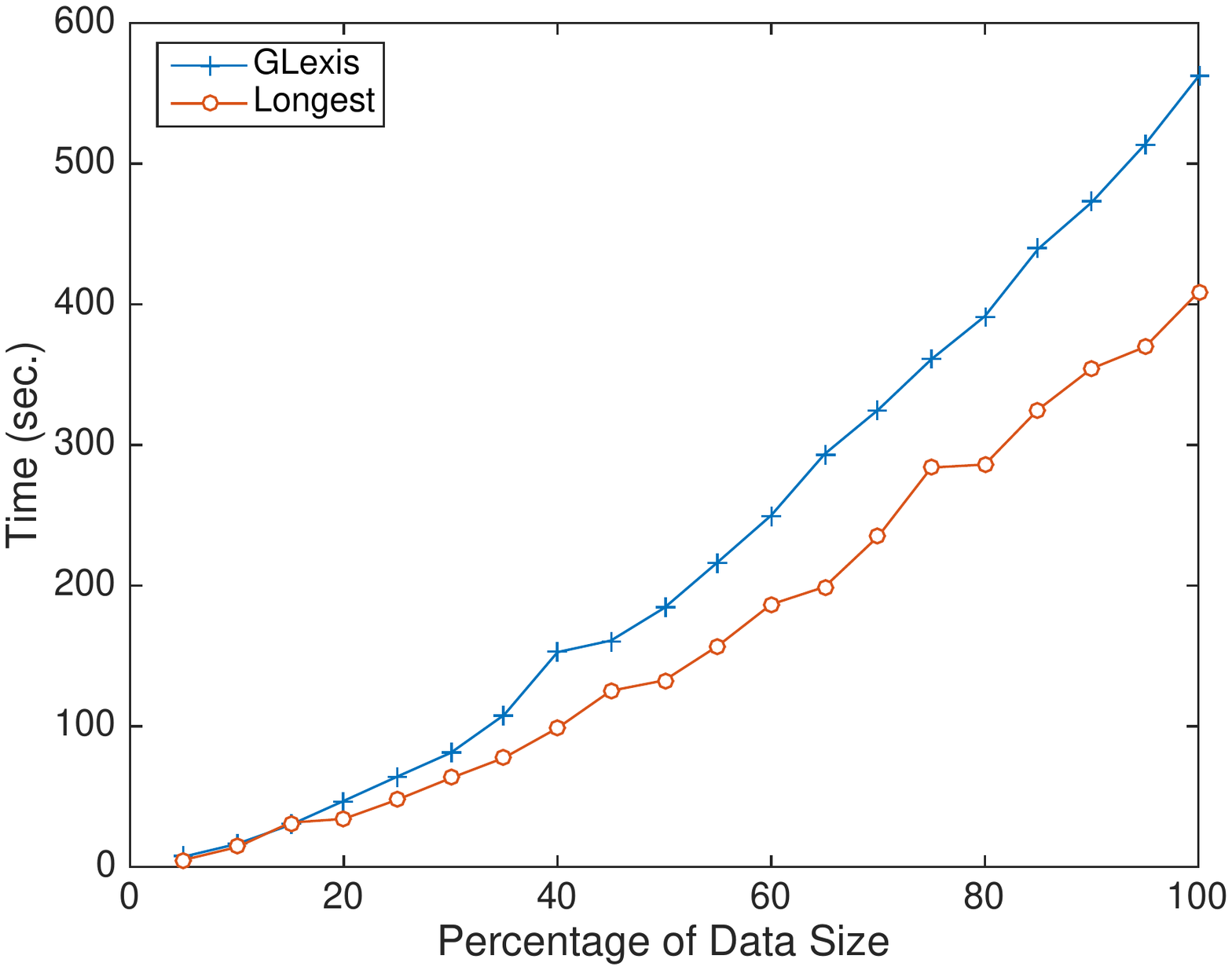}
\caption{Runtime Comparison}
\end{subfigure}
\caption{\small Comparison of \textsc{G-Lexis} with Longest Substring Replacement algorithm.
}
\label{lnrsres}
\end{figure}

We compare \textsc{G-Lexis} with an algorithm that greedily replaces the longest repeated substring, 
in terms of both runtime and cost.
We implemented the latter, originally proposed in \cite{linearlnrs} using suffix trees, using our own efficient linked-suffix array. 
We used the NSF data described in the main text and ran the two algorithms on different fractions of the total dataset, repeating the experiments
10 times and recording the average runtime and edge cost. 
As seen in Fig. \ref{lnrsres}, the Longest Substring Replacement heuristic offers a better runtime but 
its cost becomes increasingly worse than \textsc{G-Lexis} as the size of the dataset grows.
Also, \textsc{G-Lexis} is still reasonably fast on all datasets we have analyzed so far.

\subsection*{Central DNA Parts in iGEM Dataset}
\begin{table}[h]
\tiny
\centering
\caption{Top-15 nodes with highest path-centrality iGEM dataset's Lexis-DAG. Nodes shown in bold are already registered in the iGEM library.}
\begin{tabular}{l l l}
\thickhline
{\bf B0010-B0012} &&{\bf E0040-B0010-B0012}
\\
{\bf B0034-E0040-B0010-B0012} &&{\bf B0034-C0062-B0010-B0012}
\\
{\bf B0032-E0040-B0010-B0012}&&
B0034-C0062-B0010-B0012-R0062
\\
{\bf B0034-E1010-B0010-B0012}
&&
{\bf B0034-E1010}
\\
B0034-I732006-B0034-E0040-B0010-B0012    
&&
{\bf B0034-C0062}
\\
{\bf B0030-E0040-B0010-B0012}
&&
{\bf R0010-B0034}
\\
{\bf K228000-B0010-B0012}
&&
{\bf R0040-B0034}
\\
C0051-B0010-B0012 
\\\thickhline
\end{tabular}
\label{centIgem}
\end{table}


\end{document}